%% file: main.tex
\PassOptionsToPackage{unicode}{hyperref}
\PassOptionsToPackage{hyphens}{url}
\PassOptionsToPackage{dvipsnames,svgnames,x11names}{xcolor}
\documentclass[
]{article}
\input{preamble}
\hypersetup{
  pdftitle={Variable Importance Recovery in Tabular Foundation Models: A Simulation Study},
  pdfkeywords={variable importance, tabular foundation models,
interpretability, Shapley values, simulation study, gradient boosting.},
  colorlinks=true,
  linkcolor={black},
  filecolor={Maroon},
  citecolor={black},
  urlcolor={black},
  pdfcreator={LaTeX}}

\title{Variable Importance Recovery in Tabular Foundation Models: A Simulation Study\thanks{All errors are our own.}}
\author{Vasil Yasenov, PhD\thanks{Adobe Inc., Stanford University Immigration Policy Lab, and IZA Institute of Labor Economics \texttt{yasenov@stanford.edu}} \and Nicholas Adams-Cohen, PhD\thanks{Adobe Inc. \texttt{nadamscohen@adobe.com}}}
\date{November 20, 2026}
\begin{document}
\maketitle
\begin{abstract}
In machine learning, identifying which variables contribute predictive information can be as important as predictive accuracy. Tabular foundation models have matched or exceeded the predictive performance of tuned gradient boosting trees on many small- and medium-sized tabular datasets. Yet their ability to recover population-level variable importance has received little attention. We design and conduct a Monte Carlo simulation study to compare variable importance recovery across tabular foundation models and established machine learning baselines. Our framework distinguishes multiple population estimands, pairs each one with an aligned model-agnostic importance estimator, and evaluates recovery in terms of ranking, support, and relative magnitude. Our results show that tabular foundation models are generally competitive with established machine-learning baselines in recovering population variable importance, and often more accurate. We thus extend the evidence for their capabilities to a previously underexamined dimension, further strengthening their appeal.

\end{abstract}

\newpage
\section{Introduction}\label{introduction}

A recent class of predictive methods, known as tabular foundation models (TFMs), uses labeled observations as context to form predictions without task-specific parameter estimation. On many small- and medium-sized tabular datasets, these methods match or outperform the predictive accuracy of gradient-boosted decision trees (GBDTs) without task-specific hyperparameter tuning (\citeproc{ref-hollmann2025}{Hollmann et al. 2025}; \citeproc{ref-grinsztajn2026}{Grinsztajn et al. 2026}; \citeproc{ref-qu2026}{Qu et al. 2026}; \citeproc{ref-ma2025}{Ma et al. 2025}; \citeproc{ref-tabfm2026}{Kong and Das 2026}). Yet predictive accuracy is not the only consideration in many applied settings. Policymakers and other decision-makers may also need to understand and defend a model's reliance on its inputs, and in some contexts may be willing to accept lower predictive performance in exchange for greater interpretability. One important component of this objective is identifying the variables that are most predictive of an outcome and quantifying their relative importance. Whether TFMs can accurately recover such importance remains largely unknown.

Existing work applies SHAP to TFMs and develops pretrained attribution models, expanding the tools available for assessing feature contributions (\citeproc{ref-hollmann2025}{Hollmann et al. 2025}; \citeproc{ref-fonseca2026}{Fonseca and Stoyanovich 2026}; \citeproc{ref-shappfn2026}{Sena and Azevedo 2026}). Related studies evaluate how faithfully attribution methods explain a fitted model (\citeproc{ref-openxai}{Agarwal et al. 2022}; \citeproc{ref-li2023m4}{Li et al. 2023}). An attribution may faithfully describe a fitted model’s reliance on a variable without accurately reflecting that variable's predictive value in the population. These contributions advance attribution methodology, but do not establish how TFMs compare with established learners in recovering population-level variable importance.

This paper compares TFMs with established machine learning predictive baselines in their ability to recover population level variable importance. We design and conduct a Monte Carlo simulation study with known data-generating processes and evaluate recovery against population level ground truth. Our framework distinguishes four distinct population estimands from the importance of a fitted procedure and its finite-sample estimate. We pair each population target with an aligned model-agnostic importance estimator and evaluate importance recovery in terms of ranking, support, and relative magnitude. More broadly, this paper provides a framework for evaluating population variable importance recovery across AI-based prediction methods.

Our simulations show that TFMs are generally competitive with established machine-learning baselines in recovering population variable importance and often perform better. The size and consistency of these advantages depend on the definition of importance and the simulation setting, with some exceptions. The clearest gains arise when importance measures a variable's predictive value on its own: both TFMs outperform the boosted-tree baselines in identifying important variables and recovering their rankings and relative contributions. When importance measures the additional predictive value beyond the other inputs, TFMs also improve average ranking and magnitude recovery, although identification of the important variables is more often comparable. These findings complement the evidence on predictive accuracy, suggesting that strong TFM predictions need not come at the expense of reliable variable-importance recovery.  Overall, they strengthen the case for considering TFMs.

The rest of the paper is organized as follows. Section~\ref{literature-review} reviews the related literature and positions our study in context. Section~\ref{conceptual-framework} develops the conceptual framework, defining the population importance targets, their corresponding estimators, and the recovery metrics. Section~\ref{simulation-design} describes the simulation design, and Section~\ref{results} presents the results. Section~\ref{discussion} discusses implications for model choice, limitations, and directions for future research. The appendices provide implementation details, numerical results, and additional diagnostics.

\section{Literature Review}\label{literature-review}

Research on TFMs increasingly extends from prediction to explanation.
Predictive benchmarks compare TFMs with GBDTs across many datasets
(\citeproc{ref-qu2025}{Qu et al. 2025}, \citeproc{ref-qu2026}{2026};
\citeproc{ref-erickson2025}{Erickson et al. 2025}).
Interpretability studies also compare TFM explanations with classical
learners, using SHAP (\citeproc{ref-lundberg2017}{Lundberg and Lee 2017})
or attention-based importance on applied datasets and synthetic data with
known informative variables (\citeproc{ref-hollmann2025}{Hollmann et al. 2025};
\citeproc{ref-kolberg2025}{Kolberg et al. 2025}).
Other work uses in-context learning for Shapley and LOCO computation,
including conditional Shapley estimation
(\citeproc{ref-rundel2024}{Rundel et al. 2024};
\citeproc{ref-olsen2026}{Olsen and Christensen 2026}).
Pretrained attribution models estimate explanations from data without
access to a fitted predictor
(\citeproc{ref-fonseca2026}{Fonseca and Stoyanovich 2026}) or jointly
produce predictions and explanations
(\citeproc{ref-shappfn2026}{Sena and Azevedo 2026}).
Together, these approaches expand the evidence and tools for interpreting
TFMs. Evaluating population recovery additionally requires an explicit
importance target and an appropriate evaluation criterion.

Variable importance is not a single quantity: different definitions
capture different aspects of predictive information, particularly when
features are dependent. Population-level frameworks define algorithm-agnostic
importance through optimal predictive performance
(\citeproc{ref-williamsonfeng2020}{Williamson and Feng 2020};
\citeproc{ref-williamson2021}{Williamson et al. 2021},
\citeproc{ref-williamson2023}{2023}).
LOCO measures additional predictive value beyond the remaining covariates
(\citeproc{ref-lei2018}{Lei et al. 2018};
\citeproc{ref-verdinelli2024}{Verdinelli and Wasserman 2024a}),
whereas Shapley effects allocate explained variance
(\citeproc{ref-song2016}{Song, Nelson, and Staum 2016});
feature dependence affects their interpretation
(\citeproc{ref-verdinellishap2024}{Verdinelli and Wasserman 2024b}).
Work on observational and interventional Shapley explanations, removal-based
methods, and marginal versus conditional associations further clarifies
these choices (\citeproc{ref-chen2020}{Chen et al. 2020};
\citeproc{ref-covert2020}{Covert, Lundberg, and Lee 2020},
\citeproc{ref-covert2021}{2021}; \citeproc{ref-ewald2024}{Ewald et al. 2024}).
Permutation importance instead measures fitted-model reliance
(\citeproc{ref-fisher2019}{Fisher, Rudin, and Dominici 2019}), and
unrestricted permutation can force extrapolation by breaking feature
dependencies
(\citeproc{ref-hooker2021}{Hooker, Mentch, and Zhou 2021}).
These distinctions make estimator--target alignment essential when
comparing recovery across predictive procedures.

Simulation benchmarks offer complementary evaluation tools, ranging from
synthetic data with computable reference quantities to broader
assessments of attribution faithfulness
(\citeproc{ref-liu2021}{Liu et al. 2021};
\citeproc{ref-openxai}{Agarwal et al. 2022};
\citeproc{ref-li2023m4}{Li et al. 2023}).
Controlled local-attribution benchmarks also evaluate rank recovery and
noise-feature mass against analytically or numerically derived reference
attributions
(\citeproc{ref-agop2026}{Katakam 2026}).
Identifying designated informative variables, explaining a fitted model,
and recovering a population importance quantity remain distinct
evaluation objectives.

Our study combines these ideas to compare population variable-importance
recovery across TFMs, gradient-boosted trees, and elastic net.
We hold the importance procedures fixed across predictive models, match
each to its population target, and evaluate support, rank, and relative
magnitude recovery under known data-generating processes.
This design shows how model comparisons depend on the importance
definition, feature dependence, response function, and recovery criterion.

\section{Conceptual Framework}\label{conceptual-framework}

\subsection{Notation}\label{notation}

Let $X=(X_1,\ldots,X_p)$ denote a vector of covariates and $Y$ an outcome, with $(X,Y)\sim P$. We write $[p]=\{1,\ldots,p\}$, $X_{-j}$ for all covariates except $X_j$, and $X_S$ for the subvector indexed by $S\subseteq[p]$. An i.i.d. training sample of size $n$ is denoted by $\mathcal D_n=\{(X_i,Y_i)\}_{i=1}^n.$

For a prediction function $f$ based on $X_S$, define its population risk under loss $\ell$ as
$$
\mathcal R_S(f)
=
\mathbb E_P\!\left[\ell\{Y,f(X_S)\}\right].
$$
The corresponding population-optimal prediction function is
$$
f_S^*
\in
\arg\min_f \mathcal R_S(f),
$$
and we write $f^*=f_{[p]}^*$. Our simulations focus on regression, but the framework can also be adapted to classification. Under squared-error loss, $f_S^*(x_S)=\mathbb E[Y\mid X_S=x_S]$.

The data-generating process can equivalently be written as
$$
Y=f^*(X)+\varepsilon,
\qquad
\mathbb E[\varepsilon\mid X]=0.
$$
Let $\widehat f$ denote the fitted model obtained by applying the chosen prediction procedure to the training sample.

The empirical risk of a fitted model $f$ based on $X_S$ is evaluated on a sample of size $m$ from $P$, independent of the training sample. Throughout, fitted models are held fixed when taking population expectations, and empirical prediction losses are computed on the independent evaluation sample. Reusing $(X_i,Y_i)$ for the evaluation observations only in the following expression, define

$$
\widehat{\mathcal R}_S(f)
=
\frac{1}{m}
\sum_{i=1}^m
\ell\!\left\{
Y_i,
f(X_{i,S})
\right\}.
$$

For $S\subseteq[p]$,
write $\mathcal R_S^*=\mathcal R_S(f_S^*)$ for the optimal population
prediction error using $X_S$.
Let $\mathcal R_\varnothing^*$ denote the prediction error of the
optimal constant predictor. Define the predictive value of $X_S$ as
\begin{equation}\label{eq:predictive-value}
v(S)=\mathcal R_\varnothing^*-\mathcal R_S^*.
\end{equation}

We distinguish three conceptual variable importance quantities:
\begin{itemize}
  \item $\psi_j(P)$ denotes the \emph{population importance} of variable $j$, a functional of the data-generating distribution that serves as the target estimand.
  \item $\phi_j$ denotes the corresponding \emph{importance induced by fitted models under $P$}. It depends on $P$ and the fitted model or collection of fitted models used by the estimator.
  \item $\widehat\phi_j$ denotes the finite-sample estimate of $\phi_j$, computed using the independent evaluation sample and, where applicable, numerical approximation. 
\end{itemize}  

Thus, $\psi_j$ is the population target, $\phi_j$ is the importance induced by the fitted procedure when evaluated under $P$, and $\widehat\phi_j$ is its empirical estimate.

Let $\boldsymbol\psi=(\psi_1,\ldots,\psi_p)$, $
\boldsymbol\phi=(\phi_1,\ldots,\phi_p)$, and 
$\widehat{\boldsymbol\phi}=(\widehat\phi_1,\ldots,\widehat\phi_p).$
The total recovery error can then be decomposed as
$$
\boldsymbol\psi-\widehat{\boldsymbol\phi}
=
\underbrace{
\left(\boldsymbol\psi-\boldsymbol\phi\right)
}_{\text{model error}}
+
\underbrace{
\left(\boldsymbol\phi-\widehat{\boldsymbol\phi}\right)
}_{\text{estimation error}}.
$$
Model error is the discrepancy between the population importance target and the importance induced by the fitted procedure. It reflects both limitations of the prediction procedure and error from fitting on a finite training sample, rather than model misspecification alone. For example, a linear model like elastic net may fail to capture important nonlinear relationships in the population. Estimation error arises when evaluating the fitted procedure's importance using a finite independent evaluation sample and, where applicable, numerical approximation; it does not include training-sample error. Our study evaluates total recovery error without separately quantifying its model-error and estimation-error components. The next subsection defines the population importance targets we consider.

\subsection{\texorpdfstring{Population Level Importance
\emph{Estimands},
$\psi_j$}{Population Level Importance Estimands, \textbackslash psi\_j}}\label{population-level-importance-estimands-psi_jp}

Population level importance, $\psi_j(P)$, describes the predictive
information associated with $X_j$ under the joint distribution of
$(X,Y)$ (\citeproc{ref-williamson2021}{Williamson et al. 2021},
\citeproc{ref-williamson2023}{2023}). The question it answers is:
\emph{What value does $X_j$ provide for predicting $Y$ under
$P$?} The answer depends on the definition of importance. We consider one binary and three continuous importance estimands.

\subsubsection{Structural Importance}\label{structural-importance}

Structural importance records whether the population
conditional outcome function genuinely depends on $X_j$ over the
support of $P$:

\begin{equation}\label{eq:structural-importance}
\psi_j^{\mathrm{str}}
=
\begin{cases}
1, & \text{if no function of $X_{-j}$ equals $f^*(X)$ $P$-almost surely},\\
0, & \text{otherwise}.
\end{cases}
\end{equation}

It defines the active set $\{j:\psi_j^{\mathrm{str}}=1\}$ but provides
no measure of magnitude. Because the definition is restricted to the
support of $P$, variation at covariate combinations that cannot occur
does not make a variable structurally important.

\subsubsection{Marginal Importance}\label{marginal-importance}

Marginal importance measures how much predictive performance
improves when $X_j$ is used compared to a model with a constant only:

\begin{equation}\label{eq:marginal-importance}
\psi_j^{\mathrm{marg}}
=
v(\{j\})
=
\mathcal R_\varnothing^*-\mathcal R_{\{j\}}^*.
\end{equation}

Marginal importance can assign substantial importance to a variable that
is merely correlated with the underlying signal, even if it provides no
additional information after the other covariates are observed.

\subsubsection{Conditional Importance}\label{conditional-importance}

Conditional importance measures the additional predictive value
of $X_j$ after all other covariates have been observed:

\begin{equation}\label{eq:conditional-importance}
\psi_j^{\mathrm{cond}}
=
v([p])-v([p]\setminus\{j\})
=
\mathcal R_{[p]\setminus\{j\}}^*-\mathcal R_{[p]}^*
\ge 0.
\end{equation}

This is the population target associated with leave-one-covariate-out
procedures. It measures the unique contribution of $X_j$ and may
therefore be small or zero when other covariates contain redundant
predictive information.

\subsubsection{Shapley Importance}\label{shapley-importance}

Shapley importance averages the incremental predictive value of
$X_j$ over all possible subsets of the remaining features
(\citeproc{ref-williamsonfeng2020}{Williamson and Feng 2020}):

\begin{equation}\label{eq:shapley-importance}
\psi_j^{\mathrm{Shap}}
=
\sum_{S\subseteq[p]\setminus\{j\}}
\frac{|S|!(p-|S|-1)!}{p!}
\left[
v(S\cup\{j\})-v(S)
\right].
\end{equation}

Unlike conditional importance, which evaluates the contribution of
$X_j$ only after all other covariates have entered, Shapley importance
averages its contribution across every possible ordering. It therefore
distributes shared predictive information among correlated or redundant
variables. Under squared-error loss,
$v(S)=\operatorname{Var}(\mathbb E[Y\mid X_S])$,
so $\psi_j^{\mathrm{Shap}}$ corresponds to the unnormalized population
Shapley effect (\citeproc{ref-song2016}{Song et al. 2016}).

\subsubsection{Discussion}

The four definitions in Equations~\eqref{eq:structural-importance}--\eqref{eq:shapley-importance} measure distinct quantities and need not agree. Marginal importance credits correlated
proxies, conditional importance measures unique information, Shapley importance distributes shared information, and structural importance records only whether a variable is required. Even with independent covariates, interactions can produce different rankings. Under squared loss, the continuous estimands coincide only in very special cases such as additive models with independent covariates.

For a simple example, let $Z,W,U$ be independent standard normal variables, let $Q=\rho Z+\sqrt{1-\rho^2}\,U$ be a proxy with $|\rho|<1$, and let $Y=ZW+\varepsilon$ with independent mean-zero noise. Among the observed covariates $(Z,W,Q)$, only $Z$ and $W$ are structurally important, while all three marginal importance values are zero. Their conditional importance values are $1-\rho^2$, $1$, and $0$, respectively. Shapley importance can nevertheless allocate positive credit to $Q$ when $\rho\neq0$, because it helps predict the
interaction when $W$, but not $Z$, is observed. Thus, recovering the interaction's active variables and recovering marginal importance are different tasks, and assigning shared Shapley credit to a proxy is not an error.

The relevant estimand depends on the decision the analysis is intended to inform. For instance, structural importance suits settings in which the primary question is whether a variable belongs in the predictive signal at all, irrespective of the magnitude of its incremental contribution. Marginal importance is instead useful for screening variables, assessing the predictive content of a low-cost measurement on its own, or identifying variables that can serve as proxies when the full feature set is unavailable. Conditional importance is more appropriate for incremental measurement or data-acquisition decisions. Shapley importance is preferable when correlated measurements jointly encode the same construct and credit must be allocated across them. 

\subsection{\texorpdfstring{Fitted-Procedure Importance
$\phi_j$}{Fitted-Procedure Importance, \textbackslash phi\_j}}\label{fitted-procedure-importance-estimates-phi_j}

Let $\widehat f_S$ denote the model fitted on the training sample
using only $X_S$, with $\widehat f_{[p]}=\widehat f$. We define the
importance induced by these fitted models, $\phi_j$, using population
expectations. The corresponding empirical estimates, $\widehat\phi_j$,
replace these expectations with averages over an independent evaluation
sample and, where needed, numerical approximation. Empirical risks
are defined in Section~\ref{notation}.

We use model-agnostic variable importance estimators that can be applied to both TFMs and classical machine learning baselines. Tree-specific methods and scores, such as TreeSHAP, split gain, and Gini importance, are excluded because they are not applicable across all procedures.

\subsubsection{Structural Importance}\label{structural-importance-estimates}

To evaluate structural importance recovery, we rank variables using
each of the three importance estimators described below. We select
the top $k$ variables, where $k$ is the known number of structurally
active variables in the simulation, and compare the selected set with
the true active set.

\subsubsection{Marginal Refitting Importance}\label{marginal-refitting-importance}

Marginal refitting importance compares a constant-only fit with
a fit based on $X_j$ alone. The importance induced by these fitted models under $P$ is

\begin{equation}\label{eq:marginal-refitting-importance}
\phi_j^{\mathrm{marg}}
=
\mathcal R_\varnothing
  (\widehat f_\varnothing)
-
\mathcal R_{\{j\}}
  (\widehat f_{\{j\}}).
\end{equation}

Here $\widehat f_\varnothing$ is the training-sample mean. Under squared-error loss, if the fitted models converge in mean square under $P$ to their corresponding population-optimal prediction functions, then $\phi_j^{\mathrm{marg}}$ converges to
$\psi_j^{\mathrm{marg}}$.

\subsubsection{Leave-One-Covariate-Out (LOCO)}\label{loco-importance}

Leave-one-covariate-out (LOCO) measures the change in population prediction error when $X_j$ is excluded and the prediction procedure is reapplied (\citeproc{ref-lei2018}{Lei et al. 2018}; \citeproc{ref-verdinelli2024}{Verdinelli and Wasserman 2024a}). The importance induced by these fitted models under $P$ is

\begin{equation}\label{eq:loco-importance}
\phi_j^{\mathrm{LOCO}}
=
\mathcal R_{[p]\setminus\{j\}}
  (\widehat f_{[p]\setminus\{j\}})
-
\mathcal R_{[p]}(\widehat f).
\end{equation}

The corresponding population target is $\psi_j^{\mathrm{cond}}$. Unlike SHAP and SAGE, LOCO does not hold the fitted model fixed. It compares fitted models obtained with and without $X_j$. Under squared-error loss, if both fitted models converge in mean square under $P$ to their corresponding population-optimal prediction functions, then $\phi_j^{\mathrm{LOCO}}$ converges to $\psi_j^{\mathrm{cond}}$.

\subsubsection{SAGE}\label{sage-importance}

SAGE assigns Shapley values to reductions in expected
prediction loss rather than to individual predictions
(\citeproc{ref-covert2020}{Covert et al. 2020}). Throughout, SAGE refers
to its conditional version, which integrates missing features using
their distribution conditional on the observed features.
SAGE holds the full fitted model fixed; no new models are fitted for
the feature subsets. We define its projection onto $X_S$ as
$\widehat f_S^{\mathrm{proj}}(x_S)=
\mathbb E_P[\widehat f(X)\mid X_S=x_S]$.

The predictive value of this projected prediction function is

$$
v_{\widehat f}(S)
=
\mathcal R_\varnothing
  (\widehat f_\varnothing^{\mathrm{proj}})
-
\mathcal R_S
  (\widehat f_S^{\mathrm{proj}}),
$$

where $\widehat f_\varnothing^{\mathrm{proj}}=
\mathbb E_P[\widehat f(X)]$ is the corresponding constant
predictor. The SAGE importance under $P$ is

\begin{equation}\label{eq:sage-importance}
\phi_j^{\mathrm{SAGE}}
=
\sum_{S\subseteq[p]\setminus\{j\}}
\frac{|S|!(p-|S|-1)!}{p!}
\left[
v_{\widehat f}(S\cup\{j\})-v_{\widehat f}(S)
\right].
\end{equation}

Under squared-error loss, if the full fitted model converges to $f^*$
in mean square under $P$, then $\phi_j^{\mathrm{SAGE}}$ converges to
$\psi_j^{\mathrm{Shap}}$.

Its empirical estimate
$\widehat\phi_j^{\mathrm{SAGE}}$ uses the independent evaluation sample
and numerical approximation over coalitions and conditional feature
distributions.

\subsubsection{Discussion}

Marginal refitting and LOCO compare models fitted on
different feature sets; SAGE allocates the predictive
performance of the full fitted model. Their paired population targets
are marginal, conditional, and Shapley importance, respectively.

All procedures use $n$ observations for fitting and a
separate, independently generated sample to evaluate prediction losses
and variable importance. For TFMs, fitting supplies labeled context
without updating pretrained weights; LOCO removes $X_j$ from both the
context and the evaluation inputs.
For tunable procedures, we select hyperparameters once per replication
on the full feature set and reuse them when refitting restricted models,
avoiding the computational cost of a separate cross-validated search
for each feature subset. This protocol holds the selected settings fixed
across refits, but settings appropriate for the full model need not be
appropriate for every restricted prediction problem.
Appendix~\ref{appendix-model-protocol}
gives the tuning protocol.

\subsection{\texorpdfstring{Evaluating Importance
Recovery, $\psi_j - \widehat{\phi}_j$}{Evaluating Importance Recovery, \textbackslash psi\_j - \textbackslash widehat\textbackslash phi\_j}}\label{evaluating-importance-recovery}

We evaluate importance recovery through the discrepancy \(\boldsymbol\psi-\widehat{\boldsymbol\phi}\), pairing each population target with its designated estimator. These pairings match the population definition, loss function, and conditioning structure. For each Monte Carlo replicate, we evaluate recovery along three complementary dimensions: support ($F_1$ score), rank (Spearman correlation),
and magnitude (normalized RMSE). These dimensions capture different
aspects of recovery: support asks whether the right variables were
selected, rank assesses their ordering, and magnitude assesses their
relative sizes.
Throughout, $\mathcal I$ denotes the
evaluated covariates: all $p$ variables except under the restricted LOCO
protocol of Appendix~\ref{scope-decisions}. We describe each dimension
and its evaluation metric in turn.

\subsubsection{Support Recovery ($F_1$ Score)}\label{support-recovery}

Support recovery compares the top-ranked variables with a target active
set. For marginal, conditional, and Shapley importance, we count target
values strictly above zero as active:
$$
S_\psi=\{j\in\mathcal I:\psi_j>0\}.
$$
This definition includes all positive analytical target values, however
small. It does not threshold estimated scores or alter the values used
for rank and magnitude recovery.

We select the $k=|S_\psi|$ variables with the largest estimated
importance values to form $\widehat S_k$, then compute
$$
\operatorname{F_1}
=
\frac{2\,|\widehat S_k\cap S_\psi|}
{|\widehat S_k|+|S_\psi|}.
$$
When $k>0$, both sets have size $k$, so $F_1$ equals precision and
recall. The target support size is known rather than estimated. When
$S_\psi=\varnothing$, $F_1$ is undefined.

\subsubsection{Rank Recovery (Spearman Correlation)}\label{rank-recovery}

Rank recovery measures agreement in the ordering of importance values.
We include all evaluated variables except those designated as null
in $\mathcal J\subseteq\mathcal I$
(Section~\ref{covariates-and-variable-roles}), retaining variables
even when $\psi_j$ or $\widehat\phi_j$ is zero. We use Spearman
correlation:
\begin{equation}\label{eq:rank-recovery}
\operatorname{RankRec}
=
\operatorname{cor}\!\left[
\operatorname{rank}\!
\left(\widehat{\boldsymbol\phi}_{\mathcal J}\right),
\operatorname{rank}\!\left(\boldsymbol\psi_{\mathcal J}\right)
\right].
\end{equation}
Exact ties receive midranks in each vector. Small estimation differences
that break population ties can reduce correlation even when unequal
target values are correctly ordered. Rank recovery is undefined with fewer than two compared variables or when either vector is constant over $\mathcal J$.

\subsubsection{Magnitude Recovery (RMSE)}\label{magnitude-recovery}

Magnitude recovery compares relative importance using signed $L_1$
normalization over all evaluated variables $\mathcal I$, including nulls:
$$
\widetilde\phi_j
=
\frac{\widehat\phi_j}
{\sum_{k\in\mathcal I}|\widehat\phi_k|},
\qquad
\widetilde\psi_j
=
\frac{\psi_j}
{\sum_{k\in\mathcal I}|\psi_k|}.
$$
Signs are retained so negative estimates are penalized. We compute
$$
\operatorname{RMSE}
=
\sqrt{
\frac{1}{|\mathcal I|}
\sum_{j\in\mathcal I}
(\widetilde\phi_j-\widetilde\psi_j)^2
}.
$$
When defined, RMSE penalizes zero/nonzero mismatches between $\psi_j$
and $\widehat\phi_j$. It is undefined when either vector is entirely
zero; such evaluations are reported separately.

\section{Simulation Design}\label{simulation-design}

We generate repeated training samples from controlled data-generating
processes for which the population importance vector $\boldsymbol\psi$
is known analytically. This section specifies the covariate distribution and
variable roles, the response functions, the scenario grid, the sample
sizes, and the ground truth computation needed to reproduce the design.
The design varies sample size, feature count, within-pair correlation, and the response function.
Estimator settings, model protocol,
realized Monte Carlo precision and other details are deferred to
Appendix~\ref{appendix-implementation}. 

\subsection{Predictive Algorithms}\label{predictive-algorithms}

We evaluate two tabular foundation models, TabPFN-3 and TabICLv2, against
three gradient-boosted tree implementations, XGBoost, LightGBM, and
CatBoost, and an elastic-net baseline. We selected the TFMs for their prominence in
tabular prediction, reflected in their popularity and citation counts, and the
classical baselines for their widespread use in practice.
All six procedures are paired with the same
model-agnostic importance estimators rather than their native importance
scores.

Computational cost materially constrained the simulation design.
Importance estimation requires repeated model fitting or prediction
across feature subsets, making a broad Monte Carlo comparison
particularly costly for TFMs. We therefore balanced scenario coverage
against tuning depth, estimator coverage, and Monte Carlo precision:
we limited tuning, reduced the TabPFN-3 ensemble, used 50 replications
per scenario, evaluated SAGE only at $p=10$, and restricted LOCO
feature evaluation when $n=200,p=50$. These trade-offs mean that our
results compare the specified computational protocols, rather than
the best attainable performance of each model family. Furthermore, each boosted-tree procedure receives a random
search over ten hyperparameter configurations with five-fold
cross-validation, and TabPFN-3 uses a four-member ensemble rather
than the default eight. Software, model settings, and tuning protocols are given in
Appendix~\ref{appendix-model-protocol}.

\subsection{Covariates and Variable Roles}\label{covariates-and-variable-roles}

Covariates are drawn as $X\sim N(0,\Sigma)$ with unit marginal
variances, $\Sigma_{jj}=1$. Every covariate receives one of four
construction labels:

\begin{itemize}
\item \emph{signal}: a base variable that enters $f^*$, directly or
through its average with a redundant variable. These variables let us
assess recovery of genuine response drivers.
\item \emph{null}: independent of every other covariate and absent
from $f^*$. They test whether estimators avoid assigning importance to
irrelevant variables.
\item \emph{redundant}: an additional variable paired with a signal;
both enter $f^*$ through their average, so swapping them leaves
$f^*$ unchanged. They test whether estimators distinguish shared
predictive information from a variable's unique contribution.
\item \emph{proxy}: an additional variable paired with a signal but
absent from $f^*$; it is correlated with the signal when $\rho>0$.
They test whether estimators distinguish predictive information carried
through correlation from direct involvement in the response function.
\end{itemize}

The active grid starts with
$p_{\mathrm{sig}}$ base signal variables.
The first $n_{\mathrm{red}}=\lfloor p_{\mathrm{sig}}/2\rfloor$
form redundant pairs with additional covariates, and the next
$n_{\mathrm{proxy}}=\lfloor p_{\mathrm{sig}}/2\rfloor$ form
signal--proxy pairs with additional covariates. The pairs are disjoint.
Both members of a redundant pair are structurally active; designating
one as the base signal is only bookkeeping.
The remaining
$p_{\mathrm{null}}=p-p_{\mathrm{sig}}-n_{\mathrm{red}}-n_{\mathrm{proxy}}$
covariates are null.
Role assignment is independently re-randomized across the covariate
positions $[p]$ for every Monte Carlo replicate, so recovery metrics are
not confounded with covariate position.

$\Sigma$ is block-diagonal with a $2\times2$ block
$\left(\begin{smallmatrix}1&\rho\\\rho&1\end{smallmatrix}\right)$ for
each redundant pair $(j,j')$ and each signal--proxy pair $(j,k)$, and identity
elsewhere; $\rho$ controls correlation within each pair.
For a redundant pair, $f^*$ replaces the plain term in
$X_j$ with the symmetric average $(X_j+X_{j'})/2$. When the pair is
strongly correlated, dropping either variable leaves most of the
pair's predictive content in the other (small $\psi^{\mathrm{cond}}$
for both). Shapley importance splits the pair's contribution equally. LOCO and SAGE aim to recover these respective patterns.
A proxy $X_k$ never enters $f^*$, so
$\psi_k^{\mathrm{cond}}=0$ exactly, but it can have positive marginal
and Shapley importance when correlated with signal. Positive
marginal-refitting or SAGE scores for proxies can therefore be appropriate,
whereas LOCO targets zero. 

This taxonomy separates target-specific behavior:
independent nulls have no population importance, proxies can carry
standalone or shared predictive information without a unique
conditional contribution, and redundant pairs can have small
conditional importance despite positive structural and Shapley
importance. These are differences between targets, not necessarily
failures of recovery. Finite fitted estimates need not equal these
population targets.

For rank recovery, $\mathcal J$ contains all evaluated
non-null covariates, retaining proxies. Excluding nulls avoids having
the correlation dominated by separation of their zero-importance block
from the remaining variables, rather than ordering within the latter
(Appendix~\ref{scope-decisions-rank}).

\subsection{Response Functions}\label{response-functions-section}

Let $j_1,\ldots,j_{p_{\mathrm{sig}}}$ index the base signal variables (in the
order role assignment fixes them), and write $\tilde X_k$ for $X_{j_k}$,
or for $(X_{j_k}+X_{j_k'})/2$ if $j_k$ belongs to a redundant pair $(j_k,j_k')$ under
the scenario's role assignment.

We consider four response
functions:
$$
f^*_{\mathrm{lin}}(x)=\sum_{k=1}^{p_{\mathrm{sig}}}\beta_k\tilde X_k,
\qquad
f^*_{\mathrm{thr}}(x)=\sum_{k=1}^{p_{\mathrm{sig}}}\beta_k\tilde X_k
\mathbf{1}\{\tilde X_k>0\},
$$
$$
f^*_{\mathrm{nl}}(x)=\sum_{k=1}^{p_{\mathrm{sig}}}\beta_k\,
g_{(k\bmod 4)}(\tilde X_k),
\qquad
f^*_{\mathrm{int}}(x)=\sum_{m=1}^{\lfloor p_{\mathrm{sig}}/2\rfloor}
\beta_m\,\tilde X_{2m-1}\tilde X_{2m},
$$
where $g_0(x)=x$, $g_1(x)=x^2-1$, $g_2(x)=\sin(2x)$, and
$g_3(x)=|x|-\sqrt{2/\pi}$ are centered for a standard
normal argument. All scenarios use coefficients equal to 1, and a target signal-to-noise ratio of 4.
An averaged redundant pair has variance
$(1+\rho)/2$, so the same constants need not center its transformed
response; additive constants do not affect these importance targets.

We then have:
$$Y=f^*(X)+\varepsilon \text{  with  }
\varepsilon\sim N(0,\sigma^2),\quad \varepsilon\perp X$$

For the pure interaction response, $\psi_j^{\mathrm{marg}}=0$
for every variable because each term multiplies independent, centered
signal blocks. Signal variables nevertheless have positive conditional
and Shapley importance, reflecting differences between targets rather
than a failure of marginal importance estimators.
Support, rank, and normalized
magnitude recovery are therefore undefined for this marginal target.
We instead report marginal leakage,
$\operatorname{MargLeak}=
\sum_{j=1}^p|\widehat\phi_j^{\mathrm{marg}}|/\operatorname{Var}(Y)$,
for which zero is the population target and lower
values are better. This diagnostic measures total absolute deviation
from the zero marginal target relative to outcome variance. It includes
both positive and negative estimates; negative estimates occur when a
single-variable refit has higher evaluation loss than the fitted
constant. It is not a probability or a share constrained to lie below one.

\subsection{Scenario Grid}\label{scenario-grid}\label{sample-sizes-and-replication}

The grid crosses $n\in\{200,1000\}$,
$p\in\{10,50\}$, $\rho\in\{0,0.5,0.9\}$, and the four response
functions, giving 48 scenarios. Covariate-role counts are determined by $p$,
not independently crossed. For each scenario we compute the relevant structural, marginal, conditional, and Shapley population targets. All scenarios above keep $n$ and $p$ within the published context limits of the two primary TFMs (TabPFN-3 and TabICLv2), so neither model needs to subsample rows to fit within context.

We completed $B=50$ Monte Carlo replicates for all grid scenarios, reflecting the compute cost of repeated full-context TFM fits and refits.
Within each replicate, an independent evaluation sample of size
$m=2000$ is drawn fresh from $P$ and used both to compute test-sample
prediction error and to estimate importance, $\widehat{\boldsymbol\phi}$. The training sample $\mathcal D_n$ is redrawn independently each replicate. SAGE uses a random 128-point subset of this evaluation sample.

The primary evaluation metric for prediction is normalized
excess risk $(\widehat R-\sigma^2)/\operatorname{Var}(Y)$, where
$\widehat R$ is the mean squared prediction loss of the full fitted
model on this evaluation sample and $\sigma^2$ is the noise variance.
SAGE settings and Monte Carlo precision are documented in
Appendix~\ref{appendix-implementation}.

\subsection{Ground Truth Computation}\label{ground-truth-computation}

$v(S)=\operatorname{Var}(\mathbb E[Y\mid X_S])$ is computed
analytically using Gaussian conditional moments. The additive
responses decompose into independent signal-block contributions;
each interaction combines two independent blocks. Shapley additivity
therefore reduces the population game to local terms involving at
most four covariates, even at $p=50$. Exact local coalition enumeration
preserves known ties and zeros rather than ranking numerical
integration noise. Independent nulls receive zero importance under all targets; proxies have zero conditional importance.

We use 40,000 auxiliary Monte Carlo draws to estimate signal and outcome
variances for noise calibration and normalization; the importance targets
are computed analytically. Appendix~\ref{appendix-analytical-truth}
derives the Gaussian moment identities and exact local Shapley
calculation, and describes calibration and validation.

\section{Results}\label{results}

Figures~\ref{fig-structural-recovery}--\ref{fig-rank-recovery-shapley}
present structural, marginal, conditional, and Shapley importance
recovery, respectively. Each keeps the response functions separate
and averages over sample size $n$, feature count $p$, and feature
correlation $\rho$: 12 equally weighted scenario means per response,
each based on $B=50$ replications. SAGE was evaluated only at $p=10$,
so its results average six $n\times\rho$ scenario means per response.
Sections~\ref{results-structural}--\ref{results-shapley} describe these
figures in turn.

Figures~\ref{fig-recovery-heterogeneity}
and~\ref{fig-recovery-heterogeneity-heatmap} instead pool response
functions to compare each TFM with mean boosted-tree performance.
Figure~\ref{fig-recovery-heterogeneity} varies one design factor and
averages over the other two; Figure~\ref{fig-recovery-heterogeneity-heatmap}
retains each $(n,p,\rho)$ combination and averages only over responses.
Both weight the four responses equally, except for marginal recovery,
which excludes the interaction-only response because its target is
identically zero. Section~\ref{results-additional} discusses these
comparisons and the predictive-performance results.

Appendix~\ref{appendix-results-tables} provides numerical recovery
results in Tables~\ref{tbl-marginal-recovery}--\ref{tbl-shapley-recovery}
and predictive performance in Table~\ref{tbl-predictive-performance}.
Appendix~\ref{appendix-secondary-figures} provides alternative recovery
diagnostics in
Figures~\ref{fig-secondary-recovery-marginal}--\ref{fig-secondary-recovery-shapley}
and predictive-performance plots in Figure~\ref{fig-predictive-performance}.
Structural recovery is reported only in the main figures.

\subsection{Structural Importance}\label{results-structural}

Figure~\ref{fig-structural-recovery} contains four vertically stacked
panels for the linear, threshold, nonlinear, and interaction-only
responses. The horizontal axis identifies the six predictive
procedures, grouped as TFMs, boosted trees, and elastic net; the
vertical axis measures support recovery by $F_1$, with higher values
indicating better recovery. Within each procedure, circles, squares,
and triangles show results for marginal refitting, LOCO, and SAGE.
Each estimator selects the known number of structurally active
variables, and its selected set is compared with the structural
active set. The estimators appear together because structural
importance is a binary membership target, with no continuous
importance estimator or rank and magnitude measures of its own.

Mean $F_1$ scores are shown as the colored
markers and medians as black crosses. Bars are $\pm1.96$
between-scenario standard errors, describing variation across the
grid rather than within-scenario Monte Carlo precision.
Vertical scales
differ across panels; marked axis breaks omit empty ranges without
excluding procedures or truncating error bars.
SAGE's $p=10$ scope differs from the other estimators' two feature
counts, so differences between estimator series can also reflect
the scenarios included. The same conventions also apply
to Figures~\ref{fig-rank-recovery-marginal}--\ref{fig-rank-recovery-shapley}.

Overall, Figure~\ref{fig-structural-recovery} indicates broadly comparable
structural recovery for the TFMs and boosted trees.
LOCO $F_1$ averages about 0.94--0.99 for linear and threshold
responses and 0.79--0.83 for interaction-only responses.
Marginal refitting performs well outside the interaction-only design,
where its zero population target cannot identify the joint signal.
SAGE also recovers structural support well at $p=10$
(nonlinear mean $F_1$: 0.978 for TabPFN-3 and 0.984 for TabICLv2).
The elastic net is competitive for linear and threshold responses
but substantially weaker for nonlinear and interaction-only responses
(mean LOCO $F_1$: 0.517 and 0.365, respectively).

\subsection{Marginal Importance}\label{results-marginal}

Figure~\ref{fig-rank-recovery-marginal} compares marginal
refitting estimates with $\psi^{\mathrm{marg}}$. Rows show the linear,
threshold, and nonlinear responses; columns show support, rank, and
magnitude recovery. Higher $F_1$ and Spearman correlation and lower
normalized RMSE indicate better recovery. Each procedure has one
estimator series, and colors distinguish model families.

The bottom panel reports marginal
leakage, $\sum_j|\widehat\phi_j^{\mathrm{marg}}|/\operatorname{Var}(Y)$,
for the interaction-only response. Its marginal target is zero for
every variable, leaving ordinary support, rank, and normalized
magnitude recovery undefined. Lower leakage means smaller absolute
deviations from this zero target, not better identification
of the variables that jointly predict the outcome.

Under the specified refitting protocol,
Figure~\ref{fig-rank-recovery-marginal} shows that both TFMs have higher mean support and
rank recovery and lower magnitude error than each boosted tree in all
three nondegenerate responses (nonlinear mean $F_1$:
0.942 for TabPFN-3 and 0.912 for TabICLv2, versus 0.774--0.845 for
boosted trees). TabPFN-3 generally has stronger recovery
than TabICLv2, particularly in relative magnitude
(nonlinear mean RMSE: 0.014 versus 0.019, compared with 0.053--0.098
for boosted trees).
Interaction-only leakage is also lower for the TFMs, averaging about
0.07 for TabPFN-3 and 0.32 for TabICLv2, compared with 1.7--6.5 for the
boosted trees. The elastic net remains strong for linear and threshold
responses and has very low leakage, but its nonlinear recovery is
substantially poorer.

Boosted-tree single-variable refits reuse hyperparameters and tree counts
selected using all covariates, without further tuning or early stopping.
This choice can increase overfitting in restricted fits and yield negative
importance estimates. Some of the observed recovery and leakage
differences may therefore reflect the refitting protocol; separately
tuned or early-stopped restricted fits could change their magnitude.

\subsection{Conditional Importance}\label{results-conditional}

Figure~\ref{fig-rank-recovery} evaluates LOCO against
$\psi^{\mathrm{cond}}$ for all four responses, using the same layout as
Figure~\ref{fig-rank-recovery-marginal}. At $n=200,p=50$, LOCO
evaluates a reduced feature universe. Interaction-only elastic-net
rank metrics are undefined in 288 of 600 replicate evaluations and
normalized magnitude metrics in 262; its summaries use only the
defined values.

The results show higher mean rank recovery and
lower magnitude error for both TFMs than for each boosted tree in
every response class (interaction-only mean RMSE:
0.027 for TabPFN-3 and 0.028 for TabICLv2, versus 0.044--0.058 for
boosted trees). Rank differences are modest for linear and
threshold responses and larger for nonlinear and interaction-only
responses, where TFM means are about 0.81--0.82 and 0.73--0.75,
respectively (versus 0.74--0.78 and 0.65--0.71 for
boosted trees). Support recovery is more comparable, with some
boosted-tree means slightly exceeding those of the TFMs.
An exception occurs at
$n=1000,p=50,\rho=0.9$ in the interaction-only
design, conditional $F_1$ is approximately 0.67 for TabPFN-3 and
0.57 for TabICLv2, versus 0.75 for the mean boosted-tree benchmark.
The elastic net again performs well for linear and threshold
responses but much less well for nonlinear and interaction-only
responses.

\subsection{Shapley Importance}\label{results-shapley}

Figure~\ref{fig-rank-recovery-shapley} evaluates SAGE
against $\psi^{\mathrm{Shap}}$, using the same layout as
Figure~\ref{fig-rank-recovery}. Results are restricted to $p=10$
because SAGE was not evaluated at $p=50$.

Recovery is broadly similar
across the TFMs and boosted trees, rather than a consistent advantage
for one procedure on every criterion (nonlinear mean
Spearman correlation: 0.803 for TabPFN-3 and 0.812 for TabICLv2,
versus 0.788--0.816 for boosted trees). Interaction-only support
recovery is high for the TFMs, CatBoost, and LightGBM, with mean
$F_1$ around 0.92--0.96.
The clearest separation is from the elastic net, which has much
poorer nonlinear and interaction-only recovery
(mean $F_1$: 0.485 and 0.362, respectively).
Both TFMs are competitive for Shapley recovery within the evaluated
$p=10$ scope. Small differences among the stronger procedures should
be interpreted cautiously, since SAGE's finite evaluation sample,
conditional integration, and permutation sampling contribute estimation
error alongside fitted-model error.

\subsection{Additional Results}\label{results-additional}

Figure~\ref{fig-recovery-heterogeneity} presents seven rows:
structural support $F_1$, followed by support $F_1$ and Spearman rank
recovery for marginal, conditional, and Shapley importance.
Columns vary sample size, feature count, and feature correlation,
with the factor's levels on the horizontal axis.
The vertical axis is the TFM's recovery metric minus the equally
weighted mean metric for XGBoost, LightGBM, and CatBoost; positive
values favor the TFM because higher $F_1$ and Spearman
correlation indicate better recovery. Large points and lines average over responses
and the two design factors not on the horizontal axis, while faint
points show contrasts within design cells, averaged over responses.
Shapley feature-count panels show unconnected points at $p=10$,
with $p=50$ marked as not evaluated.

Figure~\ref{fig-recovery-heterogeneity} supports broad TFM
competitiveness in these pooled comparisons.
Marginal support and rank advantages are especially apparent,
and larger at $p=50$ (mean $F_1$ advantages of 0.169
for TabPFN-3 and 0.132 for TabICLv2); many structural, conditional-support, and
Shapley comparisons are closer to zero.
Small negative Shapley differences in some pooled views
(e.g., a mean $F_1$ contrast of $-0.004$ for TabICLv2 at $n=1000$) are better
read as broadly similar performance than as substantial reversals.
These averages do not establish superiority in every response,
scenario, or comparison with an individual boosted tree.

Figure~\ref{fig-recovery-heterogeneity-heatmap} retains each
$(n,p,\rho)$ combination in four target-specific blocks.
Rows identify design cells and columns identify the two TFMs.
The structural block shows LOCO support $F_1$; the other blocks show
Spearman rank recovery and normalized RMSE.
Higher $F_1$ and Spearman correlation and lower raw
RMSE indicate better recovery; larger plotted contrasts favor the TFM
for every metric.
Responses are pooled with the same weights and scope as
Figure~\ref{fig-recovery-heterogeneity}.
Support and rank contrasts subtract mean boosted-tree performance
from TFM performance; RMSE reverses the subtraction, so positive
values always favor the TFM. Blue favors the TFM and orange the
boosted-tree mean, with scales shared within each metric.
Gray SAGE cells were not evaluated, and asterisks flag the
$n=200,p=50$ restricted LOCO scope.
Neither colors nor the points in Figure~\ref{fig-recovery-heterogeneity}
indicate statistical significance.

Figure~\ref{fig-recovery-heterogeneity-heatmap} reinforces the broad
marginal-recovery advantage and generally favorable conditional
rank and magnitude comparisons, while showing more mixed Shapley
contrasts. Some marginal advantages are substantial: at
$n=200,p=50,\rho=0.5$, rank recovery exceeds the mean boosted-tree
benchmark by about 0.21 for TabPFN-3 and 0.18 for TabICLv2
(mean RMSE is lower by 0.032 and 0.023, respectively, in the same cell).
Shapley differences are generally smaller, especially at $n=1000$.
Together, Figures~\ref{fig-recovery-heterogeneity}
and~\ref{fig-recovery-heterogeneity-heatmap} locate the gains without
turning small cell-specific differences into a universal ranking.

Figures~\ref{fig-secondary-recovery-marginal}--\ref{fig-secondary-recovery-shapley}
in Appendix~\ref{appendix-secondary-figures} provide response-specific
checks using off-support mass, Kendall's $\tau_b$, and normalized
mean absolute deviation in place of $F_1$, Spearman correlation,
and RMSE. They preserve the broad marginal-recovery advantage and
more mixed conditional and Shapley comparisons, although individual
method orderings can change.
Tables~\ref{tbl-marginal-recovery}--\ref{tbl-shapley-recovery}
in Appendix~\ref{appendix-results-tables} give the numerical summaries
for both sets of metrics and the counts of undefined evaluations.

Figure~\ref{fig-predictive-performance} and
Table~\ref{tbl-predictive-performance} report predictive performance.
The figure has one panel per response function, with predictive
procedures on the horizontal axis and normalized excess risk on the
vertical axis; lower values are better. Points average the same
12 $n\times p\times\rho$ scenario means used for the non-SAGE
recovery comparisons.

Figure~\ref{fig-predictive-performance} shows that both TFMs have
lower mean normalized excess risk than each boosted
tree in every response class; the elastic net is best for the linear
response (mean normalized excess risk: 0.014, versus
0.019 for TabPFN-3 and 0.018 for TabICLv2). Interaction-only means are about 0.22 and 0.27 for the
TFMs, versus 0.40--0.55 for boosted trees.
Thus, strong predictive performance accompanies generally competitive
importance recovery, although the predictive ordering does not
transfer to every recovery criterion.

\section{Discussion}\label{discussion}

This paper asks whether tabular foundation models can compete with
established machine-learning baselines in recovering population variable
importance. Our Monte Carlo framework makes this comparison systematic
by distinguishing population targets from fitted-model estimates,
matching targets and estimators, and evaluating support, rank, and
relative magnitude recovery against known ground truth. The results
show broad competitiveness, with clearer marginal-recovery advantages
and more comparable performance on other criteria, although conditional
support recovery is lower in a highly correlated
interaction setting. Alongside their predictive performance, these
findings show that the strengths of this new generation of models
extend to an important, previously underexamined aspect of model
interpretation.

These findings do not make TFMs the preferred choice for every
application. Population variable-importance recovery does not measure
human comprehensibility or transparency of the model itself.
Requirements for an inspectable functional form, explicit coefficients,
or a sparse specification such as elastic net can favor established
models regardless of their importance-recovery performance.
Computational cost, fitting and inference speed, scalability, tuning
effort, and deployment requirements also matter, as do reproducibility,
robustness, calibration, and uncertainty quantification.
Established baselines may offer advantages on these dimensions, which
our study does not benchmark. Model choice therefore depends on the
combination of predictive performance, explanation capabilities, and
operational requirements that fits the application.

Our study has important limitations. The evidence is restricted to continuous Gaussian covariates with
controlled dependence and four regression response functions, rather
than the full diversity of applied tabular data. We evaluate two TFMs and three importance estimators under specific fitting and computational protocols, so the comparisons should not be interpreted as isolating architectural advantages or establishing optimal performance for each model family.
Reusing full-model hyperparameters may disadvantage classical baselines
when the appropriate model complexity changes with the feature set,
particularly for single-variable boosted-tree refits that retain the
full-model tree count without further early stopping. Some of the
observed TFM advantages in marginal and conditional importance recovery
may therefore reflect this protocol; our experiments do not establish
how these comparisons would change under subset-specific tuning.
SAGE uses finite evaluation data and Monte Carlo approximations with
a drift-based stopping rule, which limits the resolution of small
Shapley-recovery differences. Computational constraints also restrict
the evaluated scope.

Future work should extend the simulations to include categorical and
missing covariates, more complex dependence and interactions,
heterogeneous noise, and distribution shifts, alongside larger datasets.
Evaluating estimated conditional distributions and tuning performed separately for each feature subset would help separate model performance from explanation and
fitting protocols. Using the known covariate distribution also leaves open how recovery changes when that distribution must be estimated. A distinct extension concerns causality:
our targets describe predictive information, not effects of intervening
on variables. Explicit causal structures could support evaluation of
causally relevant variables or intervention-based quantities, but would
require corresponding causal estimands and identification assumptions.

\section*{AI Disclosure Statement}

Generative AI tools (Claude, Copilot) assisted with literature review, analytical computation of population targets, code implementation,
writing and revising the simulation design, producing an initial manuscript draft, and preparing the full supplementary appendix. The authors developed the research idea, the scientific and literature framing, the estimands, and the simulation analyses. They reviewed
and revised all AI-assisted material, tested all code, verified the reported results against stored simulation outputs, and checked each
reference against its source. The authors take full responsibility for the content, accuracy, and integrity of this paper.

\newpage

\section{References}\label{references}

\protect\phantomsection\label{refs}

\clearpage
\section{Figures}\label{figures}

\input{figures/fig-structural-recovery}
\clearpage
\input{figures/fig-rank-recovery-marginal}
\clearpage
\input{figures/fig-rank-recovery}
\clearpage
\input{figures/fig-rank-recovery-shapley}
\clearpage
\input{figures/fig-recovery-heterogeneity}
\clearpage
\input{figures/fig-recovery-heterogeneity-heatmap}

\clearpage
\appendix
\section*{Supplementary Appendix}
\section{Simulation Implementation Details}\label{appendix-implementation}

\renewcommand{\thetable}{A\arabic{table}}
\setcounter{table}{0}

This appendix collects implementation details for the simulation design
of Section~\ref{simulation-design}: estimator settings, model protocol,
the $p=50$ LOCO scope restriction, the non-null restriction for rank
recovery, realized Monte Carlo precision, aggregation across the sample-structure
grid, and analytical population targets.

\subsection{Covariate-Role Counts}\label{appendix-covariate-roles}

Table~\ref{tbl-roles} gives the role counts for each feature count.

\input{tables/tbl-roles}

\subsection{Estimator Settings}\label{appendix-estimator-settings}

\begin{itemize}
\item
  \emph{SAGE:} the permutation-sampling estimator of
  \hyperlink{cite.ref-covert2020}{Covert et al. (2020)}, with conditional
  feature removal via the Gaussian conditional law, using 128
  evaluation points and 64 conditional draws per coalition. A minimum of
  128 permutations is required. Every 32 permutations thereafter, the
  maximum coordinatewise standard deviation of the last 32 running
  estimates is compared with $5\%$ of their current cross-feature range;
  a zero range is replaced by one for this comparison. Sampling stops
  when this drift criterion holds or at 1024 permutations. This is a
  computational stopping heuristic, not a Monte Carlo standard-error or
  convergence certificate. The growing coalitions within each sampled
  permutation are scored in one batched forward pass. SAGE
  is restricted to $p=10$ for every procedure because $p=50$ was too
  costly; the other importance estimators retain both feature dimensions.
\item
  \emph{LOCO:} one full fit and $p$ leave-one-covariate-out fits per
  replicate, subject to the scope restriction below. The full fit is
  shared with prediction evaluation and SAGE.
\end{itemize}

\subsection{Model Protocol}\label{appendix-model-protocol}

\begin{itemize}
\item
  \emph{TFMs (TabPFN-3 and TabICLv2):} pinned software
  versions (\texttt{tabpfn} 8.2.0 and \texttt{tabicl} 2.1.1) and
  checkpoints fixed across the run, with software and configuration
  recorded in the durable checkpoint manifest; no task-specific
  tuning. TabPFN-3 is selected explicitly via
  \texttt{ModelVersion.V3}. To limit computation, TabPFN-3 uses
  a reduced ensemble (\texttt{n\_estimators=4}, auto-scaling disabled) in
  place of the package default of 8: a paired sensitivity check at both an
  easy and a hard $p=50$ cell found the change in importance recovery
  negligible relative to Monte Carlo error, while roughly halving the
  ensemble's compute. TabICLv2 uses its default checkpoint and eight-member
  ensemble, with its key--value cache enabled only for the full fitted
  model. TabPFN-3's fit cache is disabled.
  Both use the full training sample as context at $n\in\{200,1000\}$,
  within each model's published context limits.
\item
  \emph{Query-batch sensitivity:} TabPFN's fingerprint preprocessing can
  amplify small floating-point differences when the same query appears
  in different batches. The frozen production settings, including
  fingerprinting and batched SAGE prediction, were retained throughout
  all 50 replicates. A GPU diagnostic on one easy linear and one hard
  interaction draw at $n=200,p=10$ reproduced their saved SAGE vectors
  exactly. Holding each fitted draw and sampled coalition inputs fixed,
  separate-coalition prediction and an alternative full-coalition
  calculation changed normalized recovery RMSE by less than
  $5.1\times10^{-5}$, with unchanged support recovery and feature ranks.
  This two-draw screen does not establish general batch invariance or
  assess marginal-refit or LOCO recovery sensitivity.
\item
  \emph{XGBoost, LightGBM, CatBoost:} random search over 10
  configurations with 5-fold
  cross-validation on the full $\mathcal D_n$. Each tuning fold allows
  up to 1000 trees with early-stopping patience of 30 rounds. For the
  selected configuration, the median selected tree count across folds
  is converted to an integer and clipped to $[50,1000]$. This count
  and the other selected hyperparameters are reused for the full fit
  and all marginal-refit and LOCO restricted fits, without further
  tuning or early stopping. Tuning space is max depth
  $\in\{3,\ldots,10\}$, learning rate $\in$ log-uniform$(0.01,0.3)$,
  and subsample/column-subsample $\in(0.5,1)$. LightGBM retains the
  default maximum of 31 leaves per tree.
\item
  \emph{Elastic net:} the
  regularization path and $\alpha\in\{0.25,0.5,0.75,1\}$ plus
  $\alpha=0.001$ in place of pure ridge ($\alpha=0$), chosen by
  5-fold cross-validation on the full $\mathcal D_n$ and reused, without
  retuning, for all marginal-refit and LOCO restricted fits.
  Covariates are standardized using training-sample means and scales.
\item
  \emph{Classical software:} XGBoost 3.4.0, LightGBM 4.7.0,
  CatBoost 1.2.10, and scikit-learn 1.9.0.
\item
  \emph{Seeds:} replicate $r$ combines the master seed $10000+r$ with a
  stable hash of the semantic scenario identifier. Data generation,
  each procedure fit, and each importance estimator receive separate
  label-derived streams, so results are invariant to grid ordering,
  filtering, the requested value of $B$, and the presence or failure of
  other procedures or estimators in the TFM continuation. The GBDT and
  TFM runs are persisted separately but use the same data-generation
  and noise-calibration seeds. Within each scenario, replicate $r$
  therefore has identical training and evaluation data and the same
  feature permutation in both stores. Method comparisons are paired
  at the scenario--replicate level, although the reported summaries
  do not use paired differences.
\item
  \emph{Failed fits:} a non-convergent or errored fit is excluded from
  that procedure's results for that replicate only. Fit and estimator
  failure rates are recorded separately; affected cells are omitted
  when the corresponding failure rate exceeds $2\%$.
\end{itemize}

\subsection{Scope Restriction for \texorpdfstring{$p=50$}{p=50}
LOCO}\label{scope-decisions}

Full LOCO (all $p$ leave-one-covariate-out fits) is used for every $p=10$ scenario and for
$p=50$ scenarios at $n=1000$. For $p=50$ at $n=200$, LOCO is restricted
to the covariates with a defined structural role (signal, redundant, or
proxy — at most $p_{\mathrm{sig}}+n_{\mathrm{red}}+n_{\mathrm{proxy}}$
variables under Table~\ref{tbl-roles}) plus a fixed random sample of 10
null variables, rather than refitting on every null variable. Null
variables are exchangeable under the design, and random feature
permutation makes the first ten nulls in feature-index order a random
subset. This reduces computation but changes the scoring universe:
support selection, off-support mass, and magnitude normalization use
the evaluated covariates only, with no extrapolation to unscored nulls.
These diagnostics are not identical to their full-$p$ counterparts.
Results tables and heterogeneity figures flag the restriction.

\subsection{Scope Restriction for Rank Recovery}\label{scope-decisions-rank}

Spearman and Kendall's $\tau_b$ (Section~\ref{rank-recovery}) are
computed over non-null variables only, rather than over all $p$.
Null variables have $\psi_j=0$ exactly and represent $60\%$ of the
covariates in every scenario
(Table~\ref{tbl-roles}); a correlation computed over the full vector is
then dominated by whether the null block is pushed as a group toward
the bottom of the ranking, obscuring the ordering among non-null
covariates. Restricting
$\mathcal J$ to non-null variables removes this tie-block dominance
while retaining proxies, whose target importance can be positive or
zero depending on the estimand. Support and
magnitude recovery retain evaluated nulls and proxies, subject to
the restricted LOCO universe described above.

\subsection{Monte Carlo Reporting}\label{monte-carlo-reporting}

The completed analysis uses $B=50$ replicates per scenario. For an
evaluation metric $h_r$ defined in $D\ge2$ replicates of a scenario--procedure cell,
we report the realized Monte Carlo standard error of its mean as
\[
\widehat{\mathrm{MCSE}}(\bar h)
=\frac{s_h}{\sqrt D},\qquad
s_h^2=\frac{1}{D-1}\sum_{r:\,h_r\ \mathrm{defined}}(h_r-\bar h)^2.
\]
This is the sampling variability of the replicate mean, not the
Fisher-$z$ standard error of a correlation estimated from $B$ paired
observations. Mathematically undefined evaluation metrics are not
imputed; their defined counts and exclusion rates are retained alongside
the scenario means. Across the scenario--procedure cells, median
within-scenario MCSEs for Spearman recovery are 0.0097 for marginal
importance, 0.0031 for conditional importance, and 0.0262 for Shapley
importance; their respective maxima are 0.0476, 0.1277, and 0.1007.
Thus $B=50$ does not establish that every small method difference is
resolved. These diagnostics do not replace the between-scenario error
bars or provide paired tests across the separately persisted, data-paired
runs.

\subsection{Aggregation Across the Sample-Structure Grid}\label{aggregation-primary-grid}

Figures~\ref{fig-structural-recovery}--\ref{fig-rank-recovery-shapley}
and the numerical tables summarize scenario-level means within each
response function. Each scenario mean uses the replicates in which the
metric is defined (Section~\ref{monte-carlo-reporting}); undefined values
are not imputed, and their counts are reported separately.
For marginal refitting, LOCO, and prediction, the summaries use the
12 equally weighted $n\times p\times\rho$ scenarios per response.
SAGE uses only the six $n\times\rho$ scenarios at $p=10$.
The structural figure therefore has different scenario coverage for
SAGE than for the refitting estimators. LOCO at $n=200,p=50$ retains
the restricted scoring universe of Section~\ref{scope-decisions}.

For $Q$ scenario means, tables report their mean and
$\mathrm{SD}(\text{scenario means})/\sqrt Q$. Figures show the same
mean, $\pm1.96$ between-scenario standard-error bars, and the median
(an ``$\times$''). Equal scenario weights define the comparison;
precision weighting would change that target.

Plot limits cover all procedure means, medians, and complete
$\pm1.96$ standard-error bars with padding, rather than forcing $F_1$
to occupy the full $[0,1]$ range or correlations to occupy $[-1,1]$.
When elastic net's entire displayed interval band is separated from
those of the other procedures by a large empty gap, a marked broken
y-axis removes that gap. Each segment has its own tick scale; no
procedure or interval is clipped or omitted, and the numerical tables
are unchanged. Limits and tick scales may differ between panels, so
visual distances should be interpreted using the displayed tick values.

The bars are calculated from scenario means, without separately
propagating their within-scenario MCSEs. Their dispersion therefore
contains both design variation and finite-replication noise.
Within-scenario precision is reported separately in
Section~\ref{monte-carlo-reporting}. Because the grid is fixed rather
than sampled from a population of tabular problems, these bars are
descriptive summaries, not inferential confidence intervals for a
general method advantage.

\subsection{Analytical Computation via Gaussian Moments}\label{appendix-analytical-truth}

The regression targets are computed from
$v(S)=\operatorname{Var}(\mathbb E[f^*(X)\mid X_S])$.
Independent mean-zero outcome noise does not enter this value.
The block-Gaussian design makes the required conditional moments
available in closed form; no numerical integration is used to define
the population importance vectors.

\subsubsection{Conditional Gaussian Moments}

Let $T=\tilde X_k$ be a signal argument with variance $a$.
For a nonempty coalition $S$, Gaussian conditioning gives
\[
\begin{aligned}
\mu_S&=\mathbb E[T\mid X_S]
      =\Sigma_{TS}\Sigma_{SS}^{-1}X_S,\\
t(S)&=\operatorname{Var}(\mu_S)
      =\Sigma_{TS}\Sigma_{SS}^{-1}\Sigma_{ST},\\
T\mid X_S&\sim N(\mu_S,a-t(S)),
\end{aligned}
\]
where $\Sigma_{TS}=\operatorname{Cov}(T,X_S)$.
For $S=\varnothing$, $\mu_S=t(S)=0$.
Draw $T^{(1)}$ and $T^{(2)}$ independently from this conditional law
given the same $X_S$. Unconditionally, they are jointly centered
Gaussian with variance $a$ each and covariance $t=t(S)$.
Conditional independence and the law of total covariance imply
\[
\operatorname{Var}(\mathbb E[g(T)\mid X_S])
=\operatorname{Cov}(g(T^{(1)}),g(T^{(2)}))
\equiv C_g(a,t).
\]
Thus the transformed conditional-mean variance reduces to a bivariate
Gaussian moment.

For the bases used in Section~\ref{response-functions-section}, the
resulting covariances are
\[
\begin{aligned}
C_{\mathrm{linear}}(a,t)&=t,\\
C_{\mathrm{square}}(a,t)&=2t^2,\\
C_{\mathrm{sine}}(a,t)&=\tfrac12
 \left[e^{-4(a-t)}-e^{-4(a+t)}\right],\\
C_{\mathrm{absolute}}(a,t)&=\frac{2}{\pi}
 \left[\sqrt{a^2-t^2}+t\arcsin(t/a)-a\right],\\
C_{\mathrm{threshold}}(a,t)&=\frac{1}{2\pi}
 \left[\sqrt{a^2-t^2}+t\{\pi/2+\arcsin(t/a)\}-a\right].
\end{aligned}
\]
Here the bases are $T$, $T^2$, $\sin(2T)$, $|T|$, and
$T\mathbf 1\{T>0\}$, respectively. Additive centering constants
do not affect covariance.

The square identity follows from the Gaussian fourth moment
$\mathbb E[(T^{(1)})^2(T^{(2)})^2]=a^2+2t^2$.
For sine, the product-to-sum identity and the Gaussian characteristic
function give
$\mathbb E[\cos(2(T^{(1)}\mp T^{(2)}))]=e^{-4(a\mp t)}$.
The absolute-value moment is
\[
\mathbb E[|T^{(1)}||T^{(2)}|]
=\frac{2}{\pi}\left[\sqrt{a^2-t^2}+t\arcsin(t/a)\right];
\]
subtracting $(\mathbb E|T|)^2=2a/\pi$ gives
$C_{\mathrm{absolute}}$. Finally,
$T\mathbf 1\{T>0\}=(T+|T|)/2$. Joint sign symmetry makes the mixed
linear--absolute covariances zero, so
$C_{\mathrm{threshold}}=(t+C_{\mathrm{absolute}})/4$.
All five covariances are zero at $t=0$ and equal
$\operatorname{Var}(g(T))$ at $t=a$.

\subsubsection{Block Values and Exact Importance Targets}

For a redundant average $T=(X_j+X_{j'})/2$,
$a=(1+\rho)/2$. Observing neither, either, or both members gives
$t=0$, $a^2$, or $a$, respectively: with one member observed,
$\mathbb E[T\mid X_j]=aX_j$.
For a plain signal $T=X_j$ with proxy $X_k$, observing neither,
only the proxy, or the signal (with or without the proxy) gives
$t=0$, $\rho^2$, or $1$. A plain independent signal has $t=0$
or $1$.

An additive term $\beta_k g(\tilde X_k)$ contributes
$\beta_k^2 C_g(a_k,t_k(S))$ to $v(S)$.
For an interaction $\beta_m T_1T_2$ between independent centered
blocks, conditional independence gives
\[
\mathbb E[T_1T_2\mid X_S]=\mu_{1,S}\mu_{2,S},
\qquad
\operatorname{Var}(\mu_{1,S}\mu_{2,S})=t_1(S)t_2(S).
\]
Its coalition value is therefore $\beta_m^2t_1(S)t_2(S)$.
The additive terms, and the disjoint interaction terms, involve
independent blocks, so their coalition values sum without cross terms.

Write $L_\ell$ for the covariates in term $\ell$'s blocks, including
any proxies, and $v_\ell$ for its coalition value. Then
$v(S)=\sum_\ell v_\ell(S\cap L_\ell)$, and
\[
\psi_j^{\mathrm{marg}}=v(\{j\}),\qquad
\psi_j^{\mathrm{cond}}=v([p])-v([p]\setminus\{j\}).
\]
Shapley additivity and the dummy-player property give
$\psi_j^{\mathrm{Shap}}=\sum_{\ell:\,j\in L_\ell}
\mathrm{Sh}_j(v_\ell)$, where, for $q_\ell=|L_\ell|$,
\[
\mathrm{Sh}_j(v_\ell)=
\sum_{A\subseteq L_\ell\setminus\{j\}}
\frac{|A|!(q_\ell-|A|-1)!}{q_\ell!}
\left[v_\ell(A\cup\{j\})-v_\ell(A)\right].
\]
Each additive term has at most two players and each interaction
at most four. Enumerating these local coalitions therefore computes
exact population Shapley values without enumerating all $2^p$
coalitions or sampling permutations.

Independent nulls are dummy players for every term. A proxy has zero
conditional importance because its signal is observed in the
leave-one-out coalition. Redundant-pair symmetry gives equal importance
to both members, and every singleton interaction value is zero.
These zeros and ties are retained exactly. Structural importance is
computed separately from membership in $f^*$, not from these moments.

\subsubsection{Calibration and Validation}

The original 40,000 auxiliary draws estimate
$\widehat V_f=\widehat{\operatorname{Var}}(f^*(X))$.
The simulation uses $\sigma^2=\widehat V_f/4$ for noise calibration
and $\widehat V_Y=\widehat V_f+\sigma^2$ as its outcome-variance
normalization scale. These stored scales are retained when scoring
the saved fitted estimates; they do not define the analytical
importance targets.

Validation checks compare the analytical targets with independent
paired-conditional covariance estimates at $p=10$ for every response
function and correlation level in the active grid. Exact null and
conditional-proxy zeros and redundant-pair ties are checked at both
$p=10$ and $p=50$. The checks also verify Shapley efficiency
$\sum_j\psi_j^{\mathrm{Shap}}=v([p])$, and oracle rank and magnitude
recovery. Before scoring, each analytical vector is aligned with its
replicate's saved feature permutation and checked against the stored
roles and available target signatures. No fitted importance estimate
or prediction error is changed.

\newpage
\section{Additional Results}\label{appendix-results-tables}

\renewcommand{\thetable}{B\arabic{table}}
\setcounter{table}{0}

This appendix collects numerical results for marginal, conditional, and
Shapley importance, followed by predictive performance on the grid.
All numerical tables use $B=50$ for
every procedure.

Structural importance is a binary membership target, evaluated by
the $F_1$ score of each estimator's selected set. It is reported in
Figure~\ref{fig-structural-recovery}, without a separate table here.

Table~\ref{tbl-marginal-recovery} reports support, rank, and magnitude
recovery of marginal refitting against $\psi^{\mathrm{marg}}$ for the
nondegenerate responses, plus interaction-only marginal leakage.

\input{tables/tbl-marginal-recovery}

Table~\ref{tbl-conditional-recovery} reports support, rank, and
magnitude recovery of LOCO against
$\psi^{\mathrm{cond}}$.

\input{tables/tbl-conditional-recovery}

Table~\ref{tbl-shapley-recovery} reports support, rank, and magnitude
recovery of SAGE against $\psi^{\mathrm{Shap}}$.

\input{tables/tbl-shapley-recovery}

Table~\ref{tbl-predictive-performance} reports predictive performance
on the grid.

\input{tables/tbl-predictive-performance}

\clearpage
\section{Secondary Recovery Metrics and Figures}\label{appendix-secondary-figures}

\renewcommand{\thefigure}{C\arabic{figure}}
\setcounter{figure}{0}

We considered three alternatives to the main-text recovery metrics:
off-support mass, Kendall's $\tau_b$, and normalized mean absolute
deviation (MAD). These diagnostics use the same grid scenarios,
$B=50$ replication budget, scope restrictions, and undefined-value
handling as the main analysis. Their numerical results appear alongside
the main metrics in Appendix~\ref{appendix-results-tables}.

Off-support mass measures the proportion of absolute estimated importance
assigned outside the matched target's strictly positive support,
within the evaluated universe $\mathcal I$:
$$
\operatorname{OffSupportMass}
=
\frac{\sum_{j\in\mathcal I\setminus S_\psi}|\widehat\phi_j|}
{\sum_{j\in\mathcal I}|\widehat\phi_j|}.
$$
This is target-specific off-support mass, not necessarily mass assigned
to variables having the structural role ``null.'' Unlike $F_1$, it measures
the allocation of importance rather than the composition of a selected
set.

Kendall's $\tau_b$ is a tie-adjusted measure of concordance between
estimated and population importance rankings. Like Spearman correlation,
it is computed over the non-null variables with an available estimate,
retaining proxies (Appendix~\ref{scope-decisions-rank}).

MAD uses the same signed $L_1$-normalized importance vectors as RMSE
(Section~\ref{magnitude-recovery}), but takes absolute rather than squared
errors:
$$
\operatorname{MAD}
=
\frac{1}{|\mathcal I|}
\sum_{j\in\mathcal I}
\left|\widetilde\phi_j-\widetilde\psi_j\right|.
$$
Off-support mass and MAD are undefined when their respective
normalizing denominators are zero; such replicates are reported
separately rather than assigned an arbitrary value.

These alternatives preserve the broad qualitative conclusions of the
main recovery analysis: both TFMs outperform the boosted-tree benchmarks
in marginal recovery, whereas conditional and Shapley comparisons remain
response- and criterion-dependent. Individual method orderings are not
identical across metrics; in particular, allocating little importance
outside the target support need not imply that the selected set is
correct. The figures below display these secondary diagnostics.

The marginal interaction-only target remains identically zero, so its
otherwise undefined recovery metrics are replaced by the marginal-leakage
diagnostic. Figure~\ref{fig-predictive-performance} additionally plots
normalized excess risk, the study's primary evaluation metric for prediction.
Structural importance has no additional rank or magnitude
metrics in this analysis; its support recovery remains in
Figure~\ref{fig-structural-recovery}.

\input{figures/fig-secondary-recovery-marginal}
\input{figures/fig-secondary-recovery-conditional}
\input{figures/fig-secondary-recovery-shapley}
\input{figures/fig-predictive-performance}

\end{document}

%% file: preamble.tex
\usepackage{xcolor}
\usepackage[margin=1.5in]{geometry}
\usepackage{amsmath,amssymb}
\usepackage{iftex}
\ifPDFTeX
  \usepackage[T1]{fontenc}
  \usepackage[utf8]{inputenc}
  \usepackage{textcomp} 
\else 
  \usepackage{unicode-math} 
  \defaultfontfeatures{Scale=MatchLowercase}
  \defaultfontfeatures[\rmfamily]{Ligatures=TeX,Scale=1}
\fi
\usepackage{lmodern}
\ifPDFTeX\else
\fi
\IfFileExists{upquote.sty}{\usepackage{upquote}}{}
\IfFileExists{microtype.sty}{
  \usepackage[]{microtype}
  \UseMicrotypeSet[protrusion]{basicmath} 
}{}
\makeatletter
\@ifundefined{KOMAClassName}{
  \IfFileExists{parskip.sty}{%
    \usepackage{parskip}
  }{
    \setlength{\parindent}{0pt}
    \setlength{\parskip}{6pt plus 2pt minus 1pt}}
}{
  \KOMAoptions{parskip=half}}
\makeatother
\makeatletter
\ifx\paragraph\undefined\else
  \let\oldparagraph\paragraph
  \renewcommand{\paragraph}{
    \@ifstar
      \xxxParagraphStar
      \xxxParagraphNoStar
  }
  \newcommand{\xxxParagraphStar}[1]{\oldparagraph*{#1}\mbox{}}
  \newcommand{\xxxParagraphNoStar}[1]{\oldparagraph{#1}\mbox{}}
\fi
\ifx\subparagraph\undefined\else
  \let\oldsubparagraph\subparagraph
  \renewcommand{\subparagraph}{
    \@ifstar
      \xxxSubParagraphStar
      \xxxSubParagraphNoStar
  }
  \newcommand{\xxxSubParagraphStar}[1]{\oldsubparagraph*{#1}\mbox{}}
  \newcommand{\xxxSubParagraphNoStar}[1]{\oldsubparagraph{#1}\mbox{}}
\fi
\makeatother

\usepackage{longtable,booktabs,array}
\usepackage{threeparttable}
\usepackage{calc} 
\usepackage{etoolbox}
\makeatletter
\patchcmd\longtable{\par}{\if@noskipsec\mbox{}\fi\par}{}{}
\makeatother
\IfFileExists{footnotehyper.sty}{\usepackage{footnotehyper}}{\usepackage{footnote}}
\makesavenoteenv{longtable}
\usepackage{graphicx}
\makeatletter
\newsavebox\pandoc@box
\newcommand*\pandocbounded[1]{
  \sbox\pandoc@box{#1}%
  \Gscale@div\@tempa{\textheight}{\dimexpr\ht\pandoc@box+\dp\pandoc@box\relax}%
  \Gscale@div\@tempb{\linewidth}{\wd\pandoc@box}%
  \ifdim\@tempb\p@<\@tempa\p@\let\@tempa\@tempb\fi
  \ifdim\@tempa\p@<\p@\scalebox{\@tempa}{\usebox\pandoc@box}%
  \else\usebox{\pandoc@box}%
  \fi%
}
\def\fps@figure{htbp}
\makeatother

\NewDocumentCommand\citeproctext{}{}
\NewDocumentCommand\citeproc{mm}{%
  \begingroup\def\citeproctext{#2}\cite{#1}\endgroup}
\makeatletter
 \let\@cite@ofmt\@firstofone
 \def\@biblabel#1{}
 \def\@cite#1#2{{#1\if@tempswa , #2\fi}}
\makeatother
\newlength{\cslhangindent}
\newlength{\csllabelwidth}
\newenvironment{CSLReferences}[2] 
 {\begin{list}{}{%
  \setlength{\itemindent}{0pt}
  \setlength{\leftmargin}{0pt}
  \setlength{\parsep}{0pt}
  \ifodd #1
   \setlength{\leftmargin}{\cslhangindent}
   \setlength{\itemindent}{-1\cslhangindent}
  \fi
  \setlength{\itemsep}{#2\baselineskip}}}
 {\end{list}}
\usepackage{calc}

\makeatletter
\@ifpackageloaded{caption}{}{\usepackage{caption}}
\AtBeginDocument{%
\ifdefined\contentsname
  \renewcommand*\contentsname{Table of contents}
\else
  \newcommand\contentsname{Table of contents}
\fi
\ifdefined\listfigurename
  \renewcommand*\listfigurename{List of Figures}
\else
  \newcommand\listfigurename{List of Figures}
\fi
\ifdefined\listtablename
  \renewcommand*\listtablename{List of Tables}
\else
  \newcommand\listtablename{List of Tables}
\fi
\ifdefined\figurename
  \renewcommand*\figurename{Figure}
\else
  \newcommand\figurename{Figure}
\fi
\ifdefined\tablename
  \renewcommand*\tablename{Table}
\else
  \newcommand\tablename{Table}
\fi
}
\@ifpackageloaded{float}{}{\usepackage{float}}
\floatstyle{ruled}
\@ifundefined{c@chapter}{\newfloat{codelisting}{h}{lop}}{\newfloat{codelisting}{h}{lop}[chapter]}
\floatname{codelisting}{Listing}

\makeatother
\makeatletter
\@ifpackageloaded{caption}{}{\usepackage{caption}}
\@ifpackageloaded{subcaption}{}{\usepackage{subcaption}}
\makeatother
\usepackage{bookmark}
\IfFileExists{xurl.sty}{\usepackage{xurl}}{} 
\newcommand{\caphead}[1]{\textbf{#1}}

%% file: figures/fig-structural-recovery.tex
\begin{figure}[htbp]
\centering
\includegraphics[width=\linewidth]{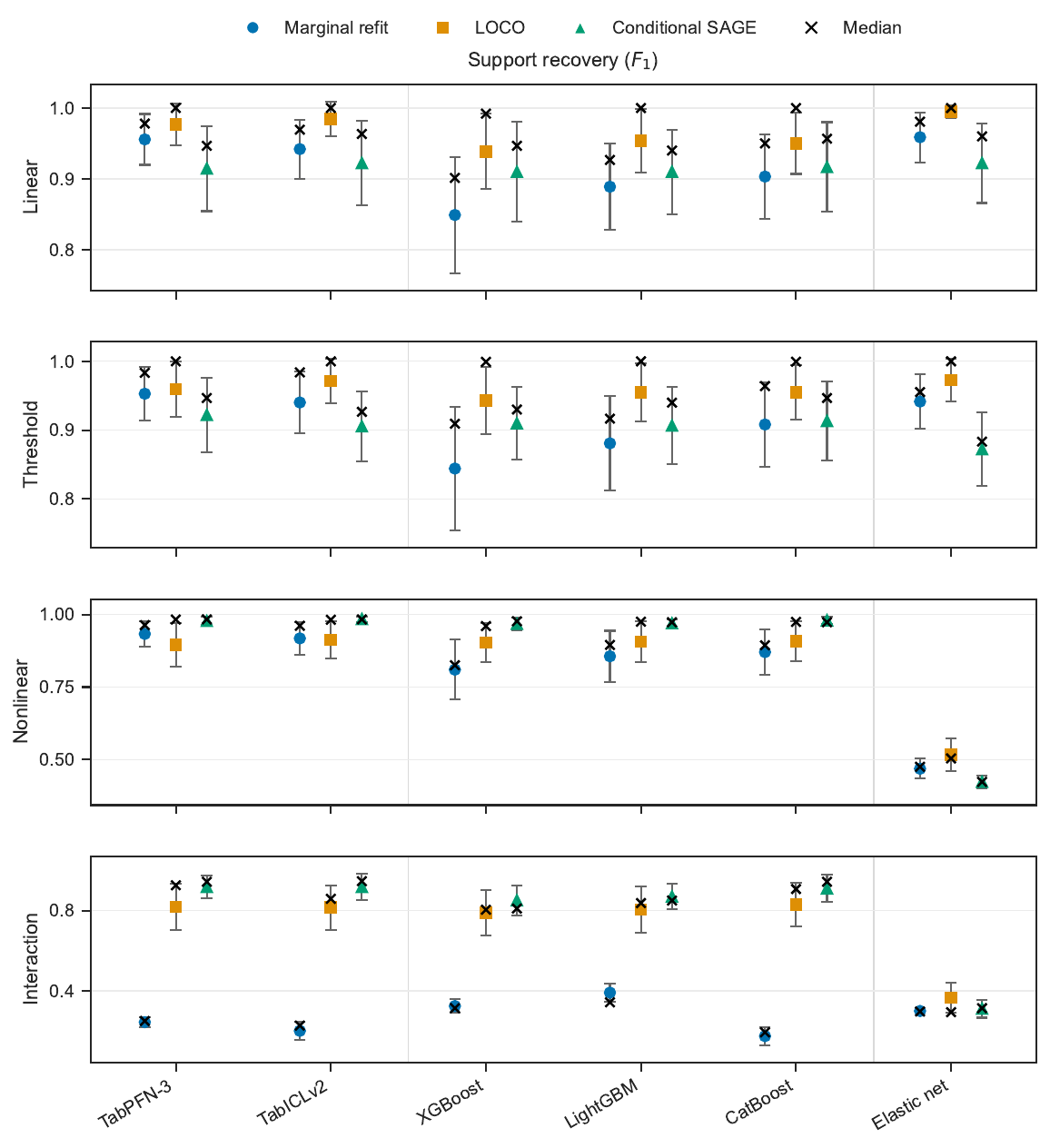}
\caption{\caphead{Structural support recovery}.
Marginal refitting and LOCO average over $n\times p\times\rho$ (12 scenarios
per response function); SAGE averages over $n\times\rho$ at
$p=10$ (six scenarios). Points are the mean $F_1$
score of each estimator's selected set $\widehat S$ against the
structural active set $S_{\psi^{\mathrm{str}}}=\{j:\psi_j^{\mathrm{str}}=1\}$;
error bars are $\pm1.96$ between-scenario standard errors across each
estimator's scenario means, and $\times$ marks their median.}\label{fig-structural-recovery}
\end{figure}

%% file: figures/fig-rank-recovery-marginal.tex
\begin{figure}[htbp]
\centering
\includegraphics[width=\linewidth]{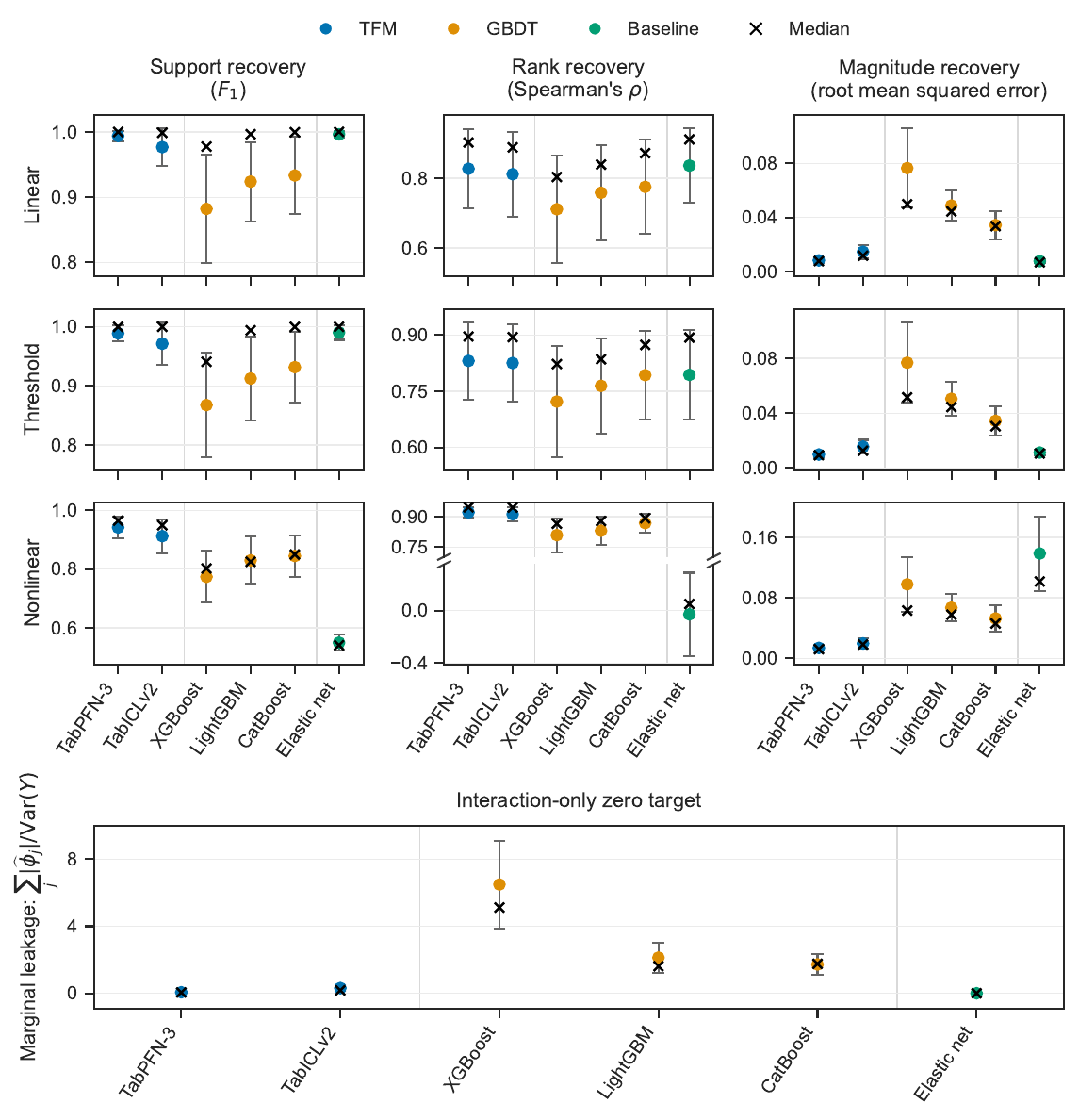}
\caption{\caphead{Marginal importance recovery}. The upper grid reports
$F_1$, Spearman $\rho$, and normalized RMSE for the three nondegenerate
response functions, averaged over the 12 $n\times p\times\rho$ scenario
means. The bottom panel reports marginal leakage
for the interaction-only response.
Points are means, error bars are $\pm1.96$ between-scenario standard
errors, and $\times$ marks the median.}\label{fig-rank-recovery-marginal}
\end{figure}

%% file: figures/fig-rank-recovery.tex
\begin{figure}[htbp]
\centering
\includegraphics[width=\linewidth]{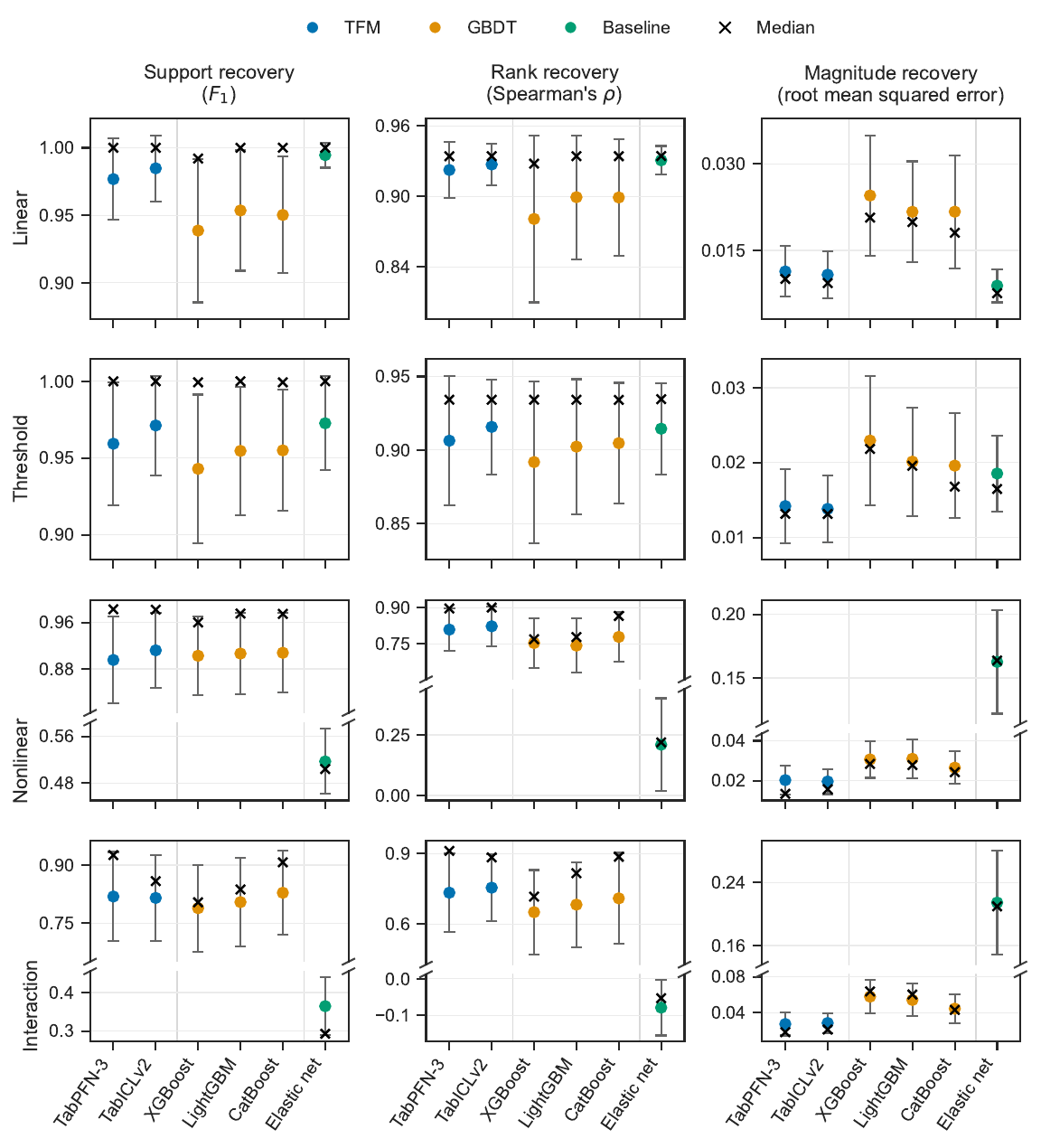}
\caption{\caphead{Conditional importance recovery}. All metrics are
averaged over the design grid ($n\times p\times\rho$). Rows are
predictive functions; columns are $F_1$ (support recovery), Spearman
$\rho$ (rank recovery), and normalized RMSE (magnitude
recovery). Error bars are
$\pm1.96$ between-scenario standard errors, and $\times$ marks the
median.}
\label{fig-rank-recovery}
\end{figure}

%% file: figures/fig-rank-recovery-shapley.tex
\begin{figure}[htbp]
\centering
\includegraphics[width=\linewidth]{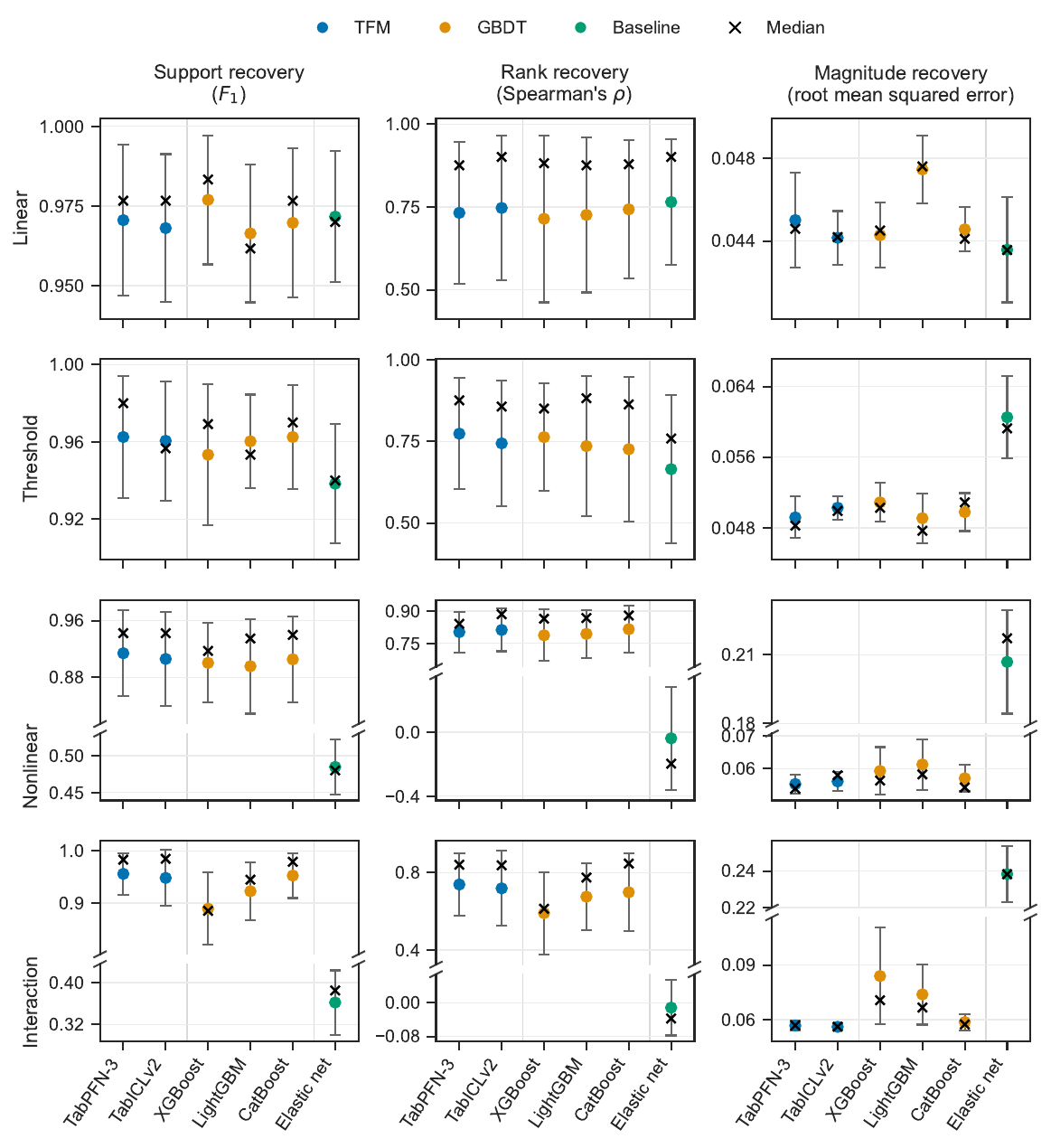}
\caption{\caphead{Shapley importance recovery}. Results restricted to $p=10$
and averaged over the six $n\times\rho$ scenario means. Rows are predictive functions; columns are
$F_1$ (support recovery), Spearman $\rho$ (rank recovery), and normalized RMSE
(magnitude recovery). Error bars are $\pm1.96$ between-scenario standard
errors, and $\times$ marks the median.}\label{fig-rank-recovery-shapley}
\end{figure}

%% file: figures/fig-recovery-heterogeneity.tex
\begin{figure}[htbp]
\centering
\includegraphics[width=\linewidth]{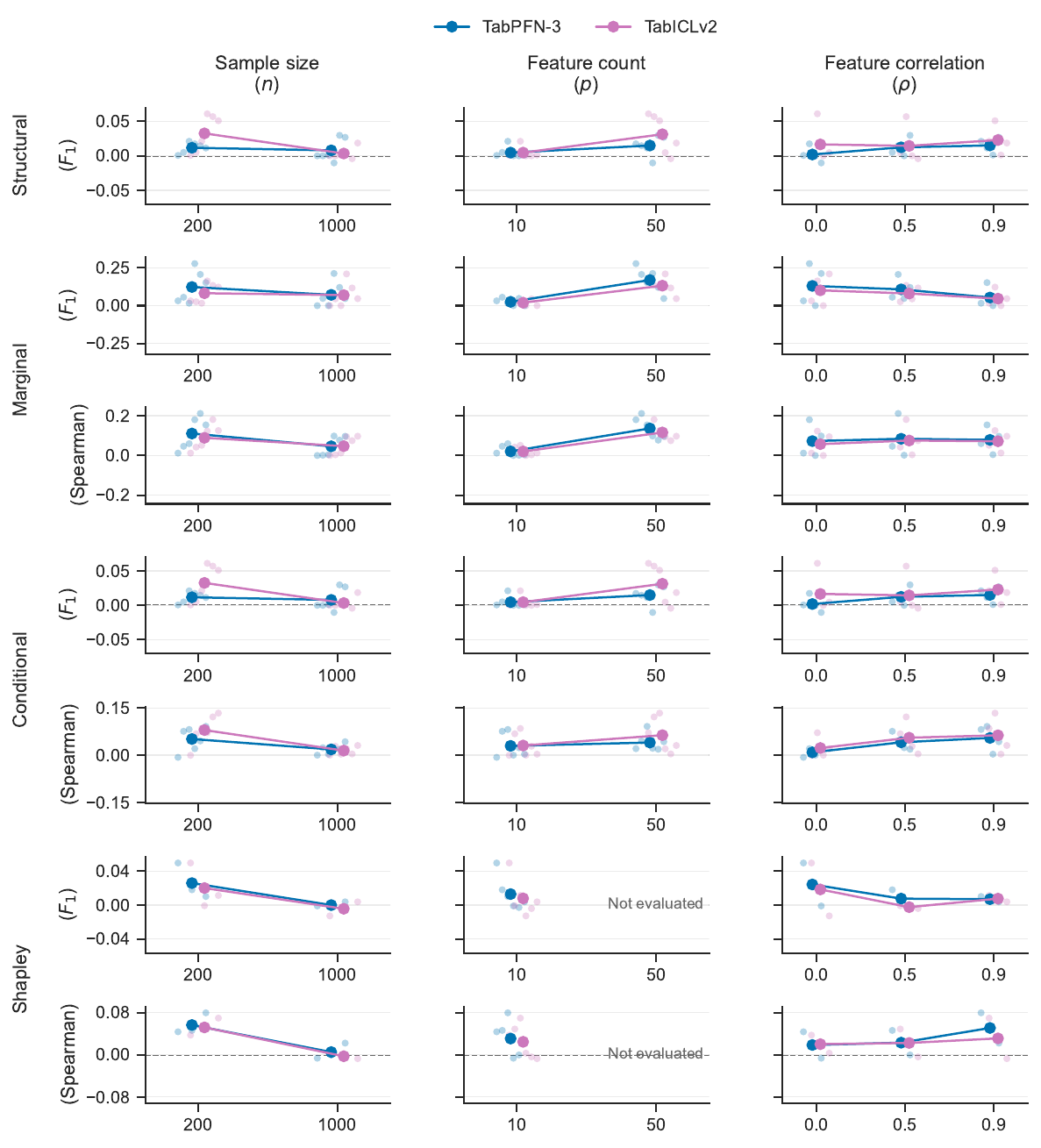}
\caption{\caphead{Importance recovery averaged over response types.}
All results are relative to mean XGBoost, LightGBM, and CatBoost performance ($B=50$).
Rows show matched-estimator support $F_1$ and Spearman rank recovery;
structural recovery uses LOCO support only. Columns vary one design
factor, averaging equally over the other two and over responses
(four except marginal's three nondegenerate responses).
Large points and lines show subgroup means; faint points show
contrasts within design cells, averaged over responses.}
\label{fig-recovery-heterogeneity}
\end{figure}

%% file: figures/fig-recovery-heterogeneity-heatmap.tex
\begin{figure}[htbp]
\centering
\includegraphics[width=\linewidth]{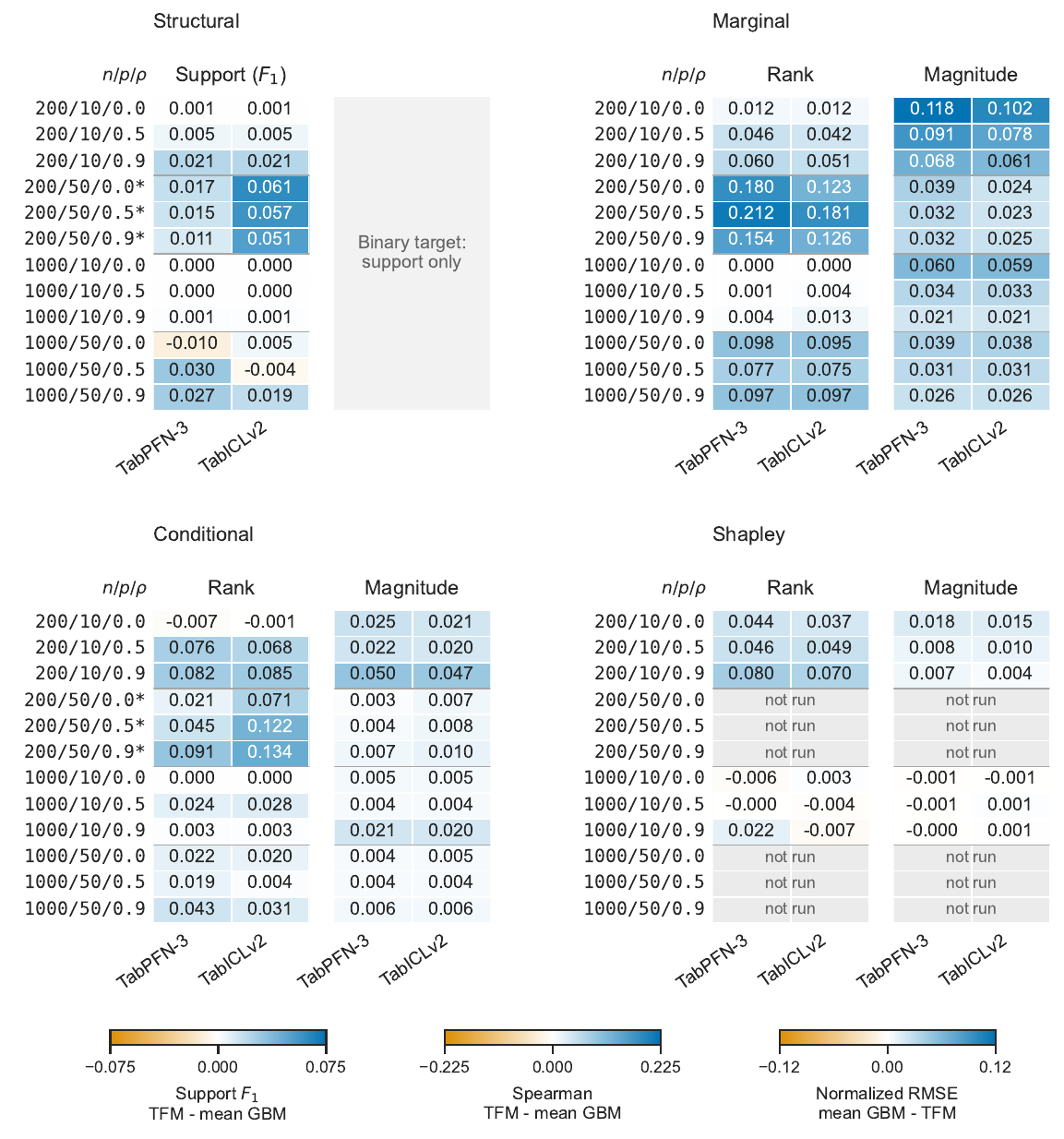}
\caption{\caphead{Importance recovery within design cells.}
All results are relative to mean XGBoost, LightGBM, and CatBoost performance ($B=50$).
Rows within each target identify $(n,p,\rho)$; columns identify the TFM.
Structural recovery is LOCO-selected support $F_1$. Rank recovery uses
Spearman correlation; magnitude recovery uses signed-$L_1$-normalized
RMSE. Support and rank differences are TFM minus mean GBM; RMSE
differences reverse the subtraction so positive values always favor
the TFM. Blue favors the TFM and orange the GBM mean. Color scales differ
by metric, not by target within a metric. Gray SAGE cells were not
evaluated; the structural target has no rank or magnitude metric.
An asterisk marks restricted LOCO scope at $n=200,p=50$.
Colors encode point estimates, not statistical significance.}
\label{fig-recovery-heterogeneity-heatmap}
\end{figure}

%% file: tables/tbl-roles.tex
\begin{table}[htbp]
\centering
\caption{\caphead{Covariate-role counts} by feature count.}\label{tbl-roles}
\begin{tabular}{@{}lcccc@{}}
\toprule
$p$ & $p_{\mathrm{sig}}$ & $n_{\mathrm{red}}$ & $n_{\mathrm{proxy}}$ & $p_{\mathrm{null}}$ \\
\midrule
10 & 2 & 1 & 1 & 6 \\
50 & 10 & 5 & 5 & 30 \\
\bottomrule
\end{tabular}
\end{table}

%% file: tables/tbl-marginal-recovery.tex
\begin{table}[htbp]
\centering
\caption{\caphead{Marginal importance recovery} on the grid. Entries are
means across 12 equally weighted scenario means within each response function, with
between-scenario standard errors in parentheses ($B=50$ per scenario).}\label{tbl-marginal-recovery}
\resizebox{\linewidth}{!}{%
\begin{tabular}{@{}lcccccc@{}}
\toprule
 & \multicolumn{2}{c}{Linear} & \multicolumn{2}{c}{Threshold} & \multicolumn{2}{c}{Nonlinear} \\
\cmidrule(lr){2-3}\cmidrule(lr){4-5}\cmidrule(lr){6-7}
\midrule
\multicolumn{7}{c}{\textit{Panel A: Support recovery}} \\
 & F1 & Off-support mass & F1 & Off-support mass & F1 & Off-support mass \\
\quad TabPFN-3 & 0.994 (0.004) & 0.031 (0.011) & 0.989 (0.007) & 0.038 (0.013) & 0.942 (0.019) & 0.044 (0.014) \\
\quad TabICLv2 & 0.977 (0.015) & 0.135 (0.044) & 0.971 (0.018) & 0.137 (0.046) & 0.912 (0.030) & 0.151 (0.046) \\
\quad XGBoost & 0.882 (0.042) & 0.603 (0.066) & 0.868 (0.045) & 0.595 (0.064) & 0.774 (0.045) & 0.624 (0.049) \\
\quad LightGBM & 0.924 (0.031) & 0.515 (0.082) & 0.912 (0.036) & 0.511 (0.078) & 0.830 (0.041) & 0.538 (0.062) \\
\quad CatBoost & 0.933 (0.030) & 0.398 (0.084) & 0.932 (0.031) & 0.381 (0.080) & 0.845 (0.036) & 0.452 (0.067) \\
\quad Elastic net & 0.996 (0.002) & 0.035 (0.011) & 0.990 (0.006) & 0.037 (0.012) & 0.549 (0.014) & 0.068 (0.014) \\
\midrule
\multicolumn{7}{c}{\textit{Panel B: Rank recovery}} \\
 & $\rho$ & $\tau_b$ & $\rho$ & $\tau_b$ & $\rho$ & $\tau_b$ \\
\quad TabPFN-3 & 0.827 (0.058) & 0.757 (0.061) & 0.831 (0.052) & 0.758 (0.057) & 0.921 (0.012) & 0.846 (0.022) \\
\quad TabICLv2 & 0.812 (0.062) & 0.742 (0.065) & 0.825 (0.052) & 0.753 (0.057) & 0.911 (0.017) & 0.835 (0.027) \\
\quad XGBoost & 0.711 (0.079) & 0.647 (0.080) & 0.722 (0.075) & 0.656 (0.077) & 0.809 (0.043) & 0.718 (0.053) \\
\quad LightGBM & 0.758 (0.070) & 0.690 (0.072) & 0.764 (0.065) & 0.693 (0.068) & 0.830 (0.036) & 0.735 (0.047) \\
\quad CatBoost & 0.775 (0.069) & 0.707 (0.071) & 0.793 (0.060) & 0.722 (0.064) & 0.868 (0.023) & 0.781 (0.035) \\
\quad Elastic net & 0.836 (0.054) & 0.765 (0.058) & 0.794 (0.060) & 0.719 (0.063) & -0.028 (0.162) & 0.011 (0.133) \\
\midrule
\multicolumn{7}{c}{\textit{Panel C: Magnitude recovery}} \\
 & RMSE & MAD & RMSE & MAD & RMSE & MAD \\
\quad TabPFN-3 & 0.008 (0.001) & 0.005 (0.001) & 0.010 (0.001) & 0.006 (0.001) & 0.014 (0.002) & 0.008 (0.001) \\
\quad TabICLv2 & 0.015 (0.003) & 0.011 (0.002) & 0.016 (0.003) & 0.011 (0.002) & 0.019 (0.004) & 0.014 (0.003) \\
\quad XGBoost & 0.076 (0.015) & 0.067 (0.013) & 0.077 (0.015) & 0.067 (0.013) & 0.098 (0.018) & 0.081 (0.016) \\
\quad LightGBM & 0.049 (0.006) & 0.043 (0.005) & 0.050 (0.006) & 0.044 (0.005) & 0.067 (0.009) & 0.056 (0.009) \\
\quad CatBoost & 0.034 (0.005) & 0.030 (0.005) & 0.034 (0.006) & 0.029 (0.005) & 0.053 (0.009) & 0.043 (0.008) \\
\quad Elastic net & 0.008 (0.001) & 0.005 (0.001) & 0.011 (0.001) & 0.006 (0.001) & 0.139 (0.025) & 0.076 (0.017) \\
\midrule
\multicolumn{7}{c}{\textit{Panel D: Interaction-only zero-target diagnostic}} \\
 & \multicolumn{6}{c}{Marginal leakage} \\
\quad TabPFN-3 & \multicolumn{6}{c}{0.066 (0.015)} \\
\quad TabICLv2 & \multicolumn{6}{c}{0.315 (0.092)} \\
\quad XGBoost & \multicolumn{6}{c}{6.491 (1.331)} \\
\quad LightGBM & \multicolumn{6}{c}{2.130 (0.453)} \\
\quad CatBoost & \multicolumn{6}{c}{1.724 (0.311)} \\
\quad Elastic net & \multicolumn{6}{c}{0.007 (0.001)} \\
\bottomrule
\end{tabular}%
}
\par\smallskip
\begin{minipage}{\linewidth}
\small\emph{Notes:} Marginal refitting is scored against $\psi^{\mathrm{marg}}$. Support uses
top-$k$ selection with $k=|S_\psi|$; off-support mass is the share of absolute estimated
importance outside that support, defined by strictly positive target values.
The interaction-only marginal target is identically
zero, so support, rank, and normalized magnitude recovery are undefined. Panel D instead
reports $\sum_j|\widehat\phi_j^{\mathrm{marg}}|/\operatorname{Var}(Y)$; lower is
better and zero is the population target.
\par An identically zero estimated vector makes off-support mass and normalized magnitude metrics undefined; a constant non-null vector also makes rank metrics undefined. Missing values are not imputed and are excluded from the corresponding means. Missing off-support mass and magnitude: 23 of 600 nonlinear replicates for Elastic net. Missing rank metrics: 23 of 600 nonlinear replicates for Elastic net.
\end{minipage}
\end{table}

%% file: tables/tbl-conditional-recovery.tex
\begin{table}[htbp]
\centering
\caption{\caphead{Conditional importance recovery} on the grid. Entries are means across
12 equally weighted scenario means within each response function, with
between-scenario standard errors in parentheses ($B=50$ per scenario).}\label{tbl-conditional-recovery}
\resizebox{\linewidth}{!}{%
\begin{tabular}{@{}lcccccccc@{}}
\toprule
 & \multicolumn{2}{c}{Linear} & \multicolumn{2}{c}{Threshold} & \multicolumn{2}{c}{Nonlinear} & \multicolumn{2}{c}{Interaction} \\
\cmidrule(lr){2-3}\cmidrule(lr){4-5}\cmidrule(lr){6-7}\cmidrule(lr){8-9}
\midrule
\multicolumn{9}{c}{\textit{Panel A: Support recovery}} \\
 & F1 & Off-support mass & F1 & Off-support mass & F1 & Off-support mass & F1 & Off-support mass \\
\quad TabPFN-3 & 0.977 (0.015) & 0.063 (0.021) & 0.959 (0.020) & 0.087 (0.027) & 0.896 (0.038) & 0.107 (0.031) & 0.819 (0.059) & 0.207 (0.058) \\
\quad TabICLv2 & 0.985 (0.013) & 0.056 (0.020) & 0.971 (0.016) & 0.085 (0.025) & 0.912 (0.033) & 0.106 (0.027) & 0.815 (0.057) & 0.219 (0.062) \\
\quad XGBoost & 0.939 (0.027) & 0.165 (0.040) & 0.943 (0.025) & 0.154 (0.035) & 0.903 (0.034) & 0.178 (0.034) & 0.789 (0.057) & 0.361 (0.054) \\
\quad LightGBM & 0.954 (0.023) & 0.144 (0.034) & 0.955 (0.021) & 0.129 (0.032) & 0.907 (0.036) & 0.170 (0.033) & 0.805 (0.058) & 0.350 (0.053) \\
\quad CatBoost & 0.950 (0.022) & 0.147 (0.037) & 0.955 (0.020) & 0.131 (0.031) & 0.908 (0.035) & 0.172 (0.034) & 0.829 (0.056) & 0.292 (0.053) \\
\quad Elastic net & 0.995 (0.005) & 0.013 (0.005) & 0.973 (0.016) & 0.031 (0.009) & 0.517 (0.028) & 0.134 (0.039) & 0.365 (0.038) & 0.521 (0.021) \\
\midrule
\multicolumn{9}{c}{\textit{Panel B: Rank recovery}} \\
 & $\rho$ & $\tau_b$ & $\rho$ & $\tau_b$ & $\rho$ & $\tau_b$ & $\rho$ & $\tau_b$ \\
\quad TabPFN-3 & 0.922 (0.012) & 0.850 (0.021) & 0.906 (0.022) & 0.834 (0.030) & 0.809 (0.045) & 0.723 (0.053) & 0.734 (0.086) & 0.671 (0.089) \\
\quad TabICLv2 & 0.927 (0.009) & 0.855 (0.019) & 0.916 (0.016) & 0.843 (0.025) & 0.823 (0.042) & 0.734 (0.050) & 0.755 (0.073) & 0.687 (0.079) \\
\quad XGBoost & 0.881 (0.036) & 0.809 (0.042) & 0.892 (0.028) & 0.819 (0.036) & 0.754 (0.053) & 0.658 (0.059) & 0.651 (0.092) & 0.592 (0.092) \\
\quad LightGBM & 0.899 (0.027) & 0.827 (0.034) & 0.902 (0.023) & 0.829 (0.031) & 0.743 (0.057) & 0.647 (0.064) & 0.683 (0.092) & 0.624 (0.093) \\
\quad CatBoost & 0.899 (0.025) & 0.826 (0.033) & 0.905 (0.021) & 0.832 (0.030) & 0.779 (0.052) & 0.686 (0.059) & 0.710 (0.099) & 0.654 (0.097) \\
\quad Elastic net & 0.931 (0.006) & 0.859 (0.017) & 0.914 (0.016) & 0.842 (0.025) & 0.210 (0.098) & 0.174 (0.086) & -0.079 (0.039) & -0.071 (0.034) \\
\midrule
\multicolumn{9}{c}{\textit{Panel C: Magnitude recovery}} \\
 & RMSE & MAD & RMSE & MAD & RMSE & MAD & RMSE & MAD \\
\quad TabPFN-3 & 0.011 (0.002) & 0.008 (0.002) & 0.014 (0.003) & 0.010 (0.002) & 0.020 (0.004) & 0.014 (0.003) & 0.027 (0.007) & 0.019 (0.005) \\
\quad TabICLv2 & 0.011 (0.002) & 0.007 (0.002) & 0.014 (0.002) & 0.010 (0.002) & 0.020 (0.003) & 0.013 (0.002) & 0.028 (0.006) & 0.020 (0.004) \\
\quad XGBoost & 0.025 (0.005) & 0.017 (0.004) & 0.023 (0.004) & 0.017 (0.003) & 0.031 (0.005) & 0.022 (0.003) & 0.058 (0.010) & 0.042 (0.007) \\
\quad LightGBM & 0.022 (0.004) & 0.015 (0.003) & 0.020 (0.004) & 0.015 (0.003) & 0.031 (0.005) & 0.022 (0.004) & 0.054 (0.009) & 0.040 (0.007) \\
\quad CatBoost & 0.022 (0.005) & 0.016 (0.004) & 0.020 (0.004) & 0.014 (0.003) & 0.027 (0.004) & 0.020 (0.003) & 0.044 (0.008) & 0.032 (0.006) \\
\quad Elastic net & 0.009 (0.001) & 0.005 (0.001) & 0.019 (0.003) & 0.010 (0.002) & 0.163 (0.021) & 0.086 (0.014) & 0.214 (0.033) & 0.117 (0.020) \\
\bottomrule
\end{tabular}%
}
\par\smallskip
\begin{minipage}{\linewidth}
\small\emph{Notes:} LOCO is scored against its matched target $\psi^{\mathrm{cond}}$. Support uses
top-$k$ selection with $k=|S_\psi|$. Off-support mass is the share of absolute estimated
importance outside the matched target's support, defined by strictly positive
target values. Rank metrics exclude structural nulls;
magnitude metrics compare signed $L_1$-normalized vectors. The $\times$ markers in the
corresponding figure show medians; table entries report means (standard errors).
At $n=200,p=50$, LOCO metrics and normalization use only evaluated covariates.
\par An identically zero estimated vector makes off-support mass and normalized magnitude metrics undefined; a constant non-null vector also makes rank metrics undefined. Missing values are not imputed and are excluded from the corresponding means. Missing off-support mass and magnitude: 262 of 600 interaction replicates for Elastic net; 23 of 600 nonlinear replicates for Elastic net. Missing rank metrics: 288 of 600 interaction replicates for Elastic net; 23 of 600 nonlinear replicates for Elastic net.
\end{minipage}
\end{table}

%% file: tables/tbl-shapley-recovery.tex
\begin{table}[htbp]
\centering
\caption{\caphead{Shapley importance recovery} on the $p=10$ grid scenarios. Entries are means across
6 equally weighted scenario means within each response function, with
between-scenario standard errors in parentheses ($B=50$ per scenario).}\label{tbl-shapley-recovery}
\resizebox{\linewidth}{!}{%
\begin{tabular}{@{}lcccccccc@{}}
\toprule
 & \multicolumn{2}{c}{Linear} & \multicolumn{2}{c}{Threshold} & \multicolumn{2}{c}{Nonlinear} & \multicolumn{2}{c}{Interaction} \\
\cmidrule(lr){2-3}\cmidrule(lr){4-5}\cmidrule(lr){6-7}\cmidrule(lr){8-9}
\midrule
\multicolumn{9}{c}{\textit{Panel A: Support recovery}} \\
 & F1 & Off-support mass & F1 & Off-support mass & F1 & Off-support mass & F1 & Off-support mass \\
\quad TabPFN-3 & 0.971 (0.012) & 0.085 (0.006) & 0.963 (0.016) & 0.087 (0.007) & 0.914 (0.031) & 0.090 (0.007) & 0.956 (0.020) & 0.117 (0.006) \\
\quad TabICLv2 & 0.968 (0.012) & 0.086 (0.007) & 0.961 (0.016) & 0.088 (0.008) & 0.906 (0.034) & 0.091 (0.008) & 0.948 (0.027) & 0.121 (0.009) \\
\quad XGBoost & 0.977 (0.010) & 0.092 (0.007) & 0.953 (0.019) & 0.095 (0.008) & 0.901 (0.029) & 0.104 (0.011) & 0.889 (0.035) & 0.190 (0.036) \\
\quad LightGBM & 0.966 (0.011) & 0.090 (0.007) & 0.960 (0.012) & 0.097 (0.009) & 0.896 (0.034) & 0.122 (0.016) & 0.923 (0.028) & 0.202 (0.037) \\
\quad CatBoost & 0.970 (0.012) & 0.089 (0.007) & 0.963 (0.014) & 0.092 (0.007) & 0.906 (0.031) & 0.097 (0.009) & 0.953 (0.022) & 0.132 (0.011) \\
\quad Elastic net & 0.972 (0.010) & 0.084 (0.007) & 0.938 (0.016) & 0.095 (0.007) & 0.485 (0.019) & 0.278 (0.024) & 0.362 (0.031) & 0.515 (0.033) \\
\midrule
\multicolumn{9}{c}{\textit{Panel B: Rank recovery}} \\
 & $\rho$ & $\tau_b$ & $\rho$ & $\tau_b$ & $\rho$ & $\tau_b$ & $\rho$ & $\tau_b$ \\
\quad TabPFN-3 & 0.733 (0.109) & 0.685 (0.112) & 0.774 (0.087) & 0.723 (0.093) & 0.803 (0.048) & 0.742 (0.054) & 0.739 (0.082) & 0.689 (0.083) \\
\quad TabICLv2 & 0.747 (0.112) & 0.707 (0.112) & 0.744 (0.098) & 0.694 (0.102) & 0.812 (0.050) & 0.756 (0.057) & 0.719 (0.099) & 0.669 (0.100) \\
\quad XGBoost & 0.715 (0.129) & 0.677 (0.126) & 0.763 (0.084) & 0.711 (0.089) & 0.788 (0.061) & 0.727 (0.070) & 0.590 (0.109) & 0.548 (0.106) \\
\quad LightGBM & 0.726 (0.119) & 0.683 (0.119) & 0.736 (0.110) & 0.694 (0.109) & 0.794 (0.057) & 0.736 (0.063) & 0.676 (0.089) & 0.621 (0.090) \\
\quad CatBoost & 0.743 (0.106) & 0.693 (0.112) & 0.726 (0.113) & 0.682 (0.114) & 0.816 (0.056) & 0.757 (0.064) & 0.699 (0.102) & 0.649 (0.104) \\
\quad Elastic net & 0.765 (0.097) & 0.716 (0.103) & 0.665 (0.115) & 0.615 (0.116) & -0.039 (0.165) & -0.027 (0.139) & -0.012 (0.033) & -0.006 (0.031) \\
\midrule
\multicolumn{9}{c}{\textit{Panel C: Magnitude recovery}} \\
 & RMSE & MAD & RMSE & MAD & RMSE & MAD & RMSE & MAD \\
\quad TabPFN-3 & 0.045 (0.001) & 0.030 (0.001) & 0.049 (0.001) & 0.033 (0.001) & 0.055 (0.002) & 0.036 (0.001) & 0.057 (0.001) & 0.039 (0.001) \\
\quad TabICLv2 & 0.044 (0.001) & 0.030 (0.000) & 0.050 (0.001) & 0.034 (0.001) & 0.056 (0.002) & 0.036 (0.001) & 0.056 (0.001) & 0.039 (0.001) \\
\quad XGBoost & 0.044 (0.001) & 0.031 (0.001) & 0.051 (0.001) & 0.034 (0.001) & 0.059 (0.004) & 0.039 (0.002) & 0.084 (0.014) & 0.057 (0.008) \\
\quad LightGBM & 0.047 (0.001) & 0.032 (0.001) & 0.049 (0.001) & 0.034 (0.001) & 0.061 (0.004) & 0.041 (0.003) & 0.074 (0.008) & 0.053 (0.006) \\
\quad CatBoost & 0.045 (0.001) & 0.031 (0.000) & 0.050 (0.001) & 0.034 (0.001) & 0.057 (0.002) & 0.038 (0.001) & 0.059 (0.002) & 0.041 (0.002) \\
\quad Elastic net & 0.044 (0.001) & 0.030 (0.001) & 0.061 (0.002) & 0.039 (0.001) & 0.207 (0.011) & 0.136 (0.008) & 0.238 (0.008) & 0.163 (0.003) \\
\bottomrule
\end{tabular}%
}
\par\smallskip
\begin{minipage}{\linewidth}
\small\emph{Notes:} SAGE is scored against its matched target $\psi^{\mathrm{Shap}}$. Support uses
top-$k$ selection with $k=|S_\psi|$. Off-support mass is the share of absolute estimated
importance outside the matched target's support, defined by strictly positive
target values. Rank metrics exclude structural nulls;
magnitude metrics compare signed $L_1$-normalized vectors. The $\times$ markers in the
corresponding figure show medians; table entries report means (standard errors).

\par An identically zero estimated vector makes off-support mass and normalized magnitude metrics undefined; a constant non-null vector also makes rank metrics undefined. Missing values are not imputed and are excluded from the corresponding means. Missing off-support mass and magnitude: 34 of 300 interaction replicates for Elastic net; 7 of 300 nonlinear replicates for Elastic net. Missing rank metrics: 34 of 300 interaction replicates for Elastic net; 7 of 300 nonlinear replicates for Elastic net.
\end{minipage}
\end{table}

%% file: tables/tbl-predictive-performance.tex
\begin{table}[htbp]
\centering
\caption{\caphead{Predictive performance} on the grid. Entries are mean
normalized excess risk across 12 equally weighted scenario means within each response
function, with between-scenario standard errors in parentheses
($B=50$ per scenario).}\label{tbl-predictive-performance}
\begin{threeparttable}
\begin{tabular}{@{}lcccc@{}}
\toprule
 & Linear & Threshold & Nonlinear & Interaction \\
\midrule
\quad TabPFN-3 & 0.019 (0.006) & 0.065 (0.021) & 0.116 (0.038) & 0.217 (0.079) \\
\quad TabICLv2 & 0.018 (0.006) & 0.065 (0.020) & 0.130 (0.039) & 0.265 (0.086) \\
\quad XGBoost & 0.120 (0.027) & 0.135 (0.031) & 0.231 (0.044) & 0.550 (0.080) \\
\quad LightGBM & 0.112 (0.023) & 0.129 (0.026) & 0.276 (0.049) & 0.524 (0.082) \\
\quad CatBoost & 0.094 (0.025) & 0.105 (0.027) & 0.200 (0.043) & 0.401 (0.079) \\
\quad Elastic net & 0.014 (0.004) & 0.243 (0.010) & 0.714 (0.023) & 0.808 (0.003) \\
\bottomrule
\end{tabular}
\begin{tablenotes}
\small
\item Normalized excess risk is
$(\widehat R-\sigma^2)/\operatorname{Var}(Y)$ on the independent evaluation sample;
lower is better and zero corresponds to the oracle risk. Normalization prevents
high-variance scenarios from receiving greater implicit weight when results are
aggregated across $n$, $p$, and $\rho$.
\end{tablenotes}
\end{threeparttable}
\end{table}

%% file: figures/fig-secondary-recovery-marginal.tex
\begin{figure}[htbp]
\centering
\includegraphics[width=\linewidth]{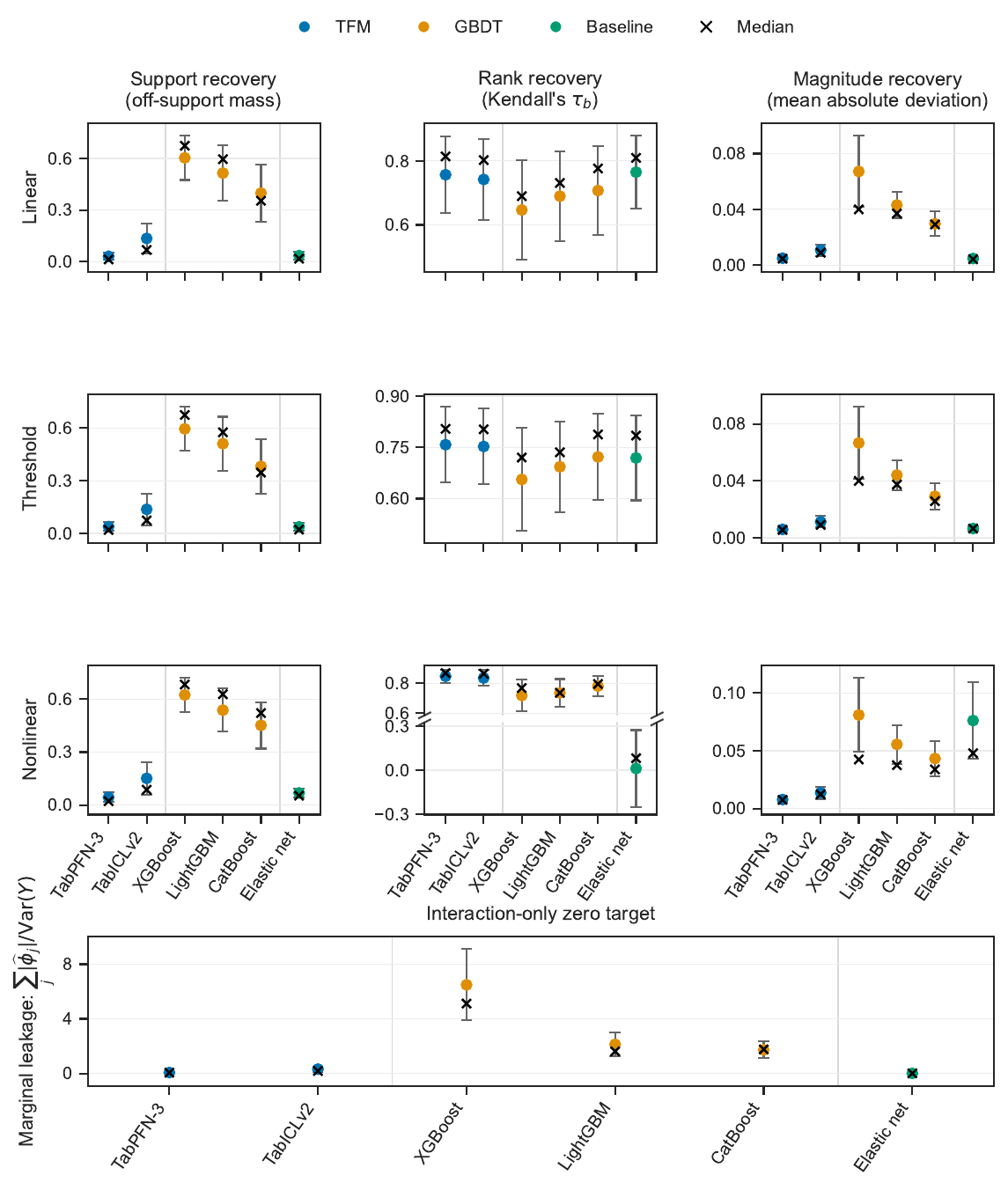}
\caption{\caphead{Marginal importance recovery: secondary metrics}
for marginal refitting against $\psi_j^{\mathrm{marg}}$. The upper grid
reports off-support mass, Kendall's $\tau_b$, and signed $L_1$-normalized
MAD, averaged over 12 equally weighted scenario means per response.
Lower off-support mass and MAD and higher $\tau_b$ indicate better recovery.
The interaction-only target is zero, so the bottom panel repeats the
marginal-leakage diagnostic rather than undefined recovery metrics.
Points are means, bars are $\pm1.96$ between-scenario standard errors,
and $\times$ marks medians. Panel-specific y limits and marked breaks
retain all methods and complete error bars ($B=50$).}
\label{fig-secondary-recovery-marginal}
\end{figure}

%% file: figures/fig-secondary-recovery-conditional.tex
\begin{figure}[htbp]
\centering
\includegraphics[width=\linewidth]{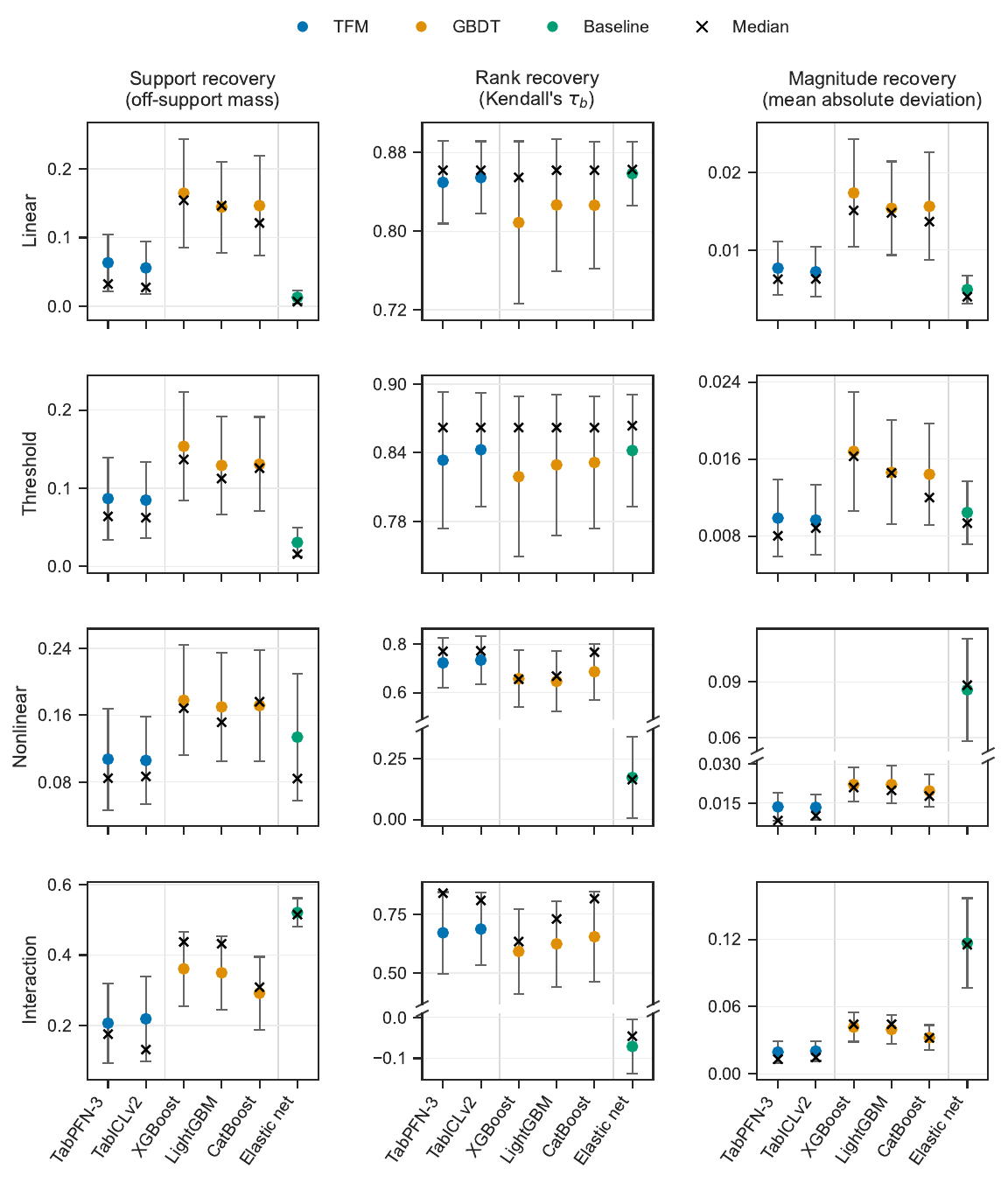}
\caption{\caphead{Conditional importance recovery: secondary metrics}
for LOCO against $\psi_j^{\mathrm{cond}}$. Rows are response functions;
columns are off-support mass, Kendall's $\tau_b$, and signed
$L_1$-normalized MAD. Lower off-support mass and MAD and higher $\tau_b$
indicate better recovery. Points average 12 equally weighted scenario
means, bars are $\pm1.96$ between-scenario standard errors, and $\times$
marks medians. The $n=200,p=50$ restricted-refit scope and undefined-value
counts are as in Table~\ref{tbl-conditional-recovery}. Panel-specific
y limits and marked breaks retain all methods and complete error bars
($B=50$).}
\label{fig-secondary-recovery-conditional}
\end{figure}

%% file: figures/fig-secondary-recovery-shapley.tex
\begin{figure}[htbp]
\centering
\includegraphics[width=\linewidth]{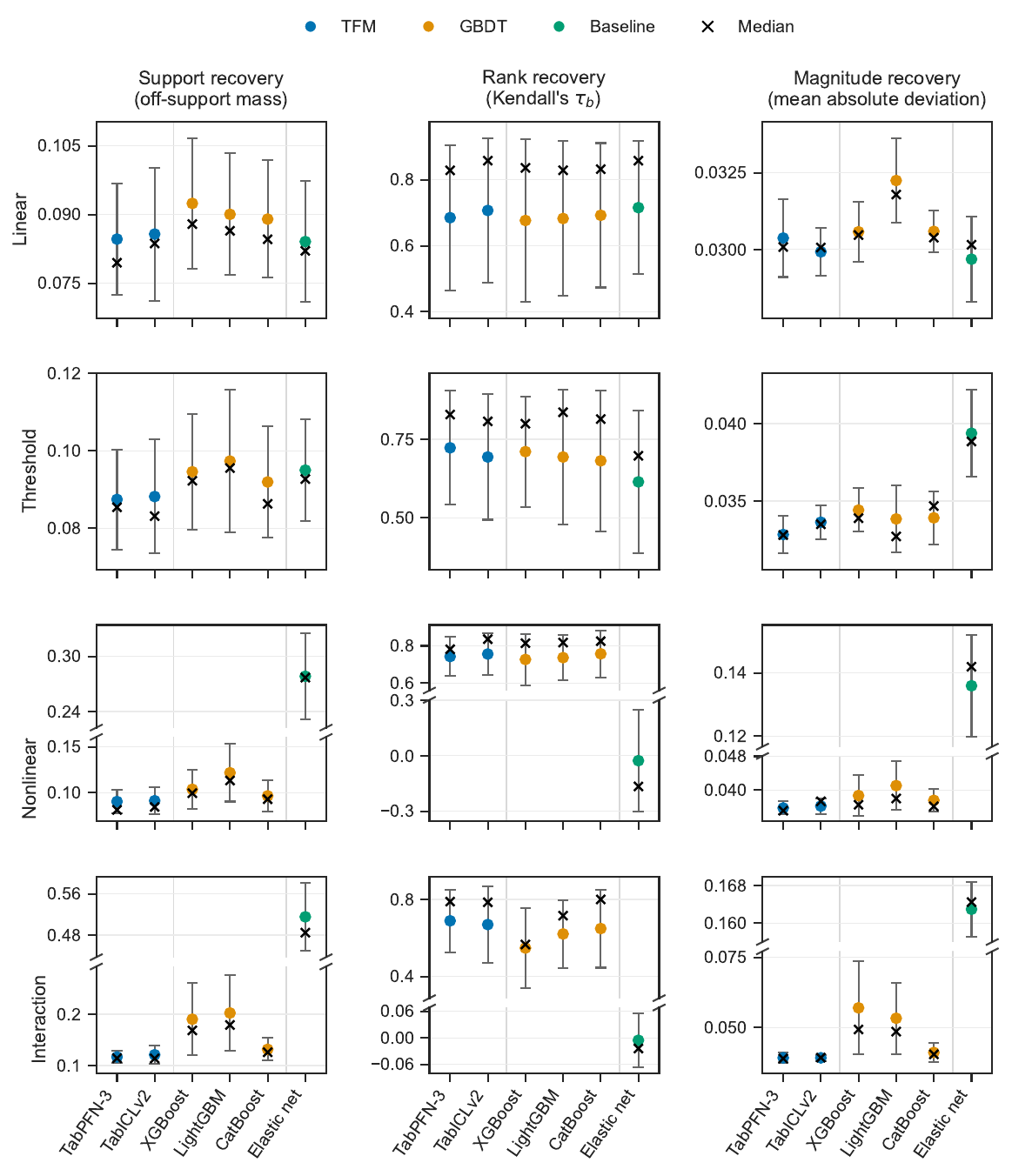}
\caption{\caphead{Shapley importance recovery: secondary metrics}
for SAGE against $\psi_j^{\mathrm{Shap}}$, restricted to
$p=10$ for every procedure. Rows are response functions; columns are
off-support mass, Kendall's $\tau_b$, and signed $L_1$-normalized MAD.
Lower off-support mass and MAD and higher $\tau_b$ indicate better recovery.
Points average six equally weighted scenario means, bars are
$\pm1.96$ between-scenario standard errors, and $\times$ marks medians.
Undefined-value counts are as in Table~\ref{tbl-shapley-recovery}.
Panel-specific y limits and marked breaks retain all methods and complete
error bars ($B=50$).}
\label{fig-secondary-recovery-shapley}
\end{figure}

%% file: figures/fig-predictive-performance.tex
\begin{figure}[htbp]
\centering
\includegraphics[width=\linewidth]{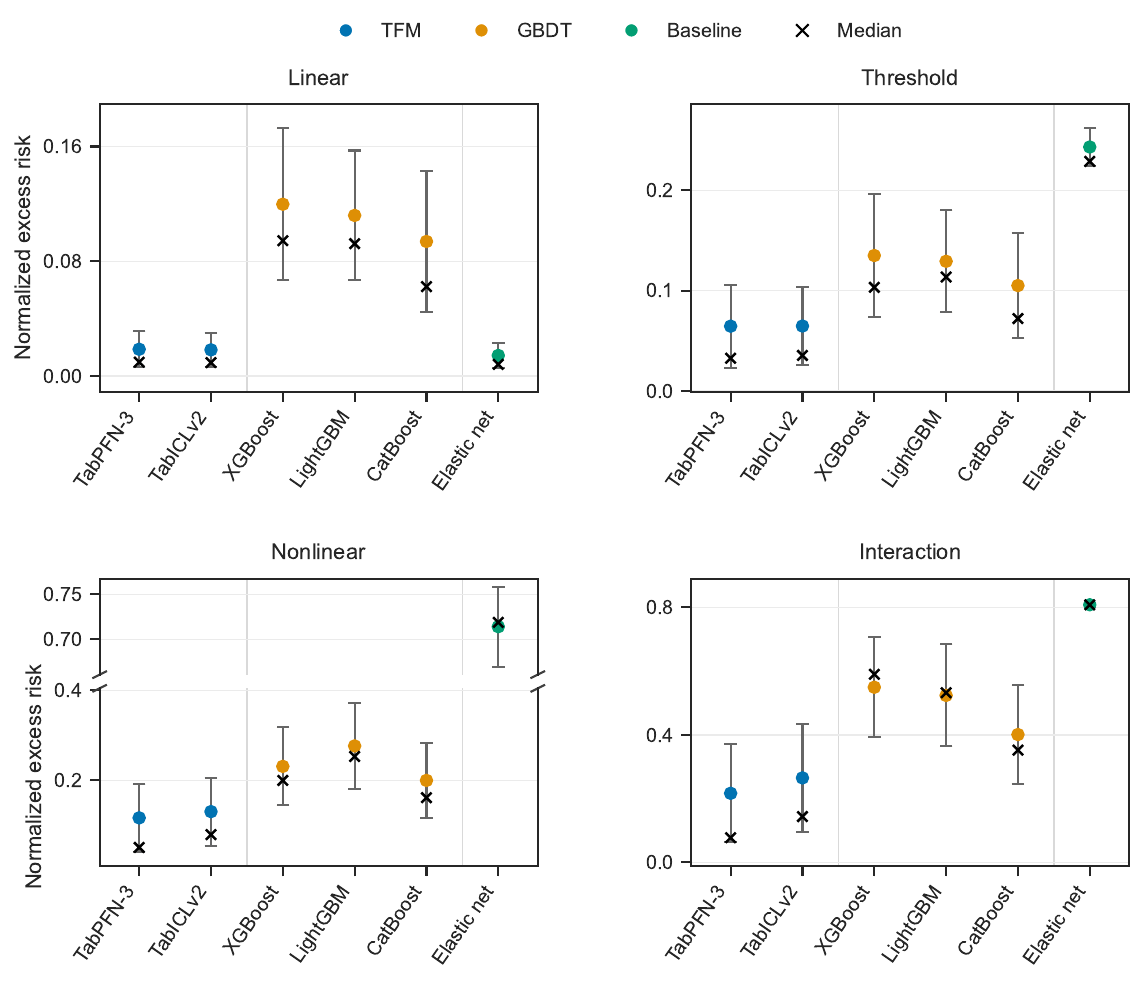}
\caption{\caphead{Predictive performance} on the grid.
Each panel reports normalized excess risk
$(\widehat R-\sigma^2)/\operatorname{Var}(Y)$ for one response function,
using the same summaries as Table~\ref{tbl-predictive-performance}.
Lower is better. Points average 12 equally weighted scenario means,
bars are $\pm1.96$ between-scenario standard errors, and $\times$
marks medians. Panel-specific y limits and marked breaks retain all
methods and complete error bars ($B=50$).}
\label{fig-predictive-performance}
\end{figure}